\documentclass{article}

\PassOptionsToPackage{numbers,compress}{natbib}

\usepackage[preprint]{neurips_2026}

\usepackage[utf8]{inputenc} 
\usepackage[T1]{fontenc}    
\usepackage{hyperref}       
\usepackage{url}            
\usepackage{booktabs}       
\usepackage{amsfonts}       
\usepackage{amsmath}        
\usepackage{amssymb}        
\usepackage{graphicx}       
\usepackage{nicefrac}       
\usepackage{microtype}      
\usepackage{xcolor}         

\usepackage{multicol}
\usepackage{multirow}
\usepackage{array}
\usepackage{colortbl}

\newcolumntype{L}[1]{>{\raggedright\arraybackslash}m{#1}}
\newcolumntype{C}[1]{>{\centering\arraybackslash}m{#1}}

\title{OptiWorld: Optimal Control for Video World Generation under Physical Constraints}

\author{
  Yu Yuan\\
  Purdue University \\
  \And
  Jianhao Yuan \\
  University of Oxford \\
  \And
  Xijun Wang \\
  Purdue University \\
 \And
  Daiqing Li \\
  SixteenMiles Labs\\
  \And
  Liu He \\
  Purdue University \\
 \And
  Lu Ling \\
  Purdue University \\
  \And
  Stanley H. Chan \\
  Purdue University \\
}

\begin{document}

\maketitle

\begin{figure}[h!]
    \centering
    \includegraphics[width=1\linewidth]{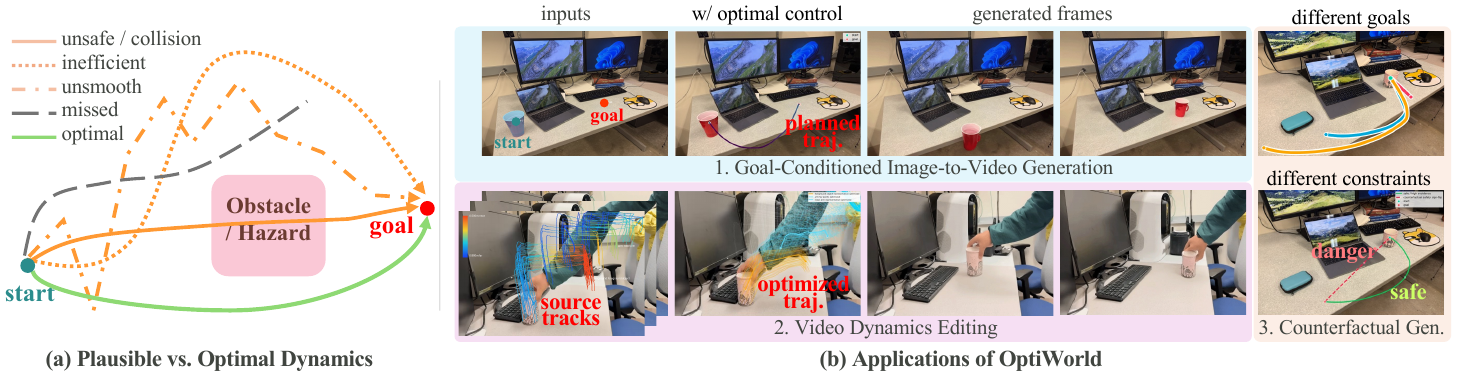}
    \caption{\textbf{Left:} without optimal control, a video generator may produce motions that look plausible but are unsafe, not smooth, or inefficient. \textbf{Right:} OptiWorld addresses these failures by introducing optimal control at video inference time: it plans a physically preferable trajectory before rendering. Zoom in for details of the planned/optimized trajectory.}
    \label{fig:teaser}
\end{figure}

\begin{abstract}
Video generation models are becoming a scalable form of world models, but they mainly generate plausible motion rather than proactively control or optimize the underlying dynamics. As a result, an object in the generated video may follow trajectories that are unsafe, not smooth, inefficient, or physically inconsistent. In this work, we propose \textbf{OptiWorld}, a framework that brings classical optimal control into video generation at inference time. OptiWorld first extracts a compact, task-relevant world state, then plans an optimal trajectory under physical constraints, and finally renders the video conditioned on this trajectory. We formulate planning as a geometric problem on a continuous manifold, which converts 3D geometry and task-dependent physical constraints into a unified planning geometry. By adding this optimal-control layer, OptiWorld generates videos with preferable dynamics, demonstrating strong potential in multiple tasks including goal-conditioned image-to-video generation, video dynamics editing, and counterfactual generation. All data and code will be available at \href{https://yuyuanspace.com/OptiWorld/}{\textcolor{blue}{https://yuyuanspace.com/OptiWorld/}}.
\end{abstract}

\section{Introduction}

World models predict future world states from current observations and control inputs \citep{Ha_2018_World, Huang_2025_towards}. Recent video generation models offer a scalable pixel-space form of world modeling: given an initial frame and a condition (often specified by prompts or actions), they can synthesize realistic future frames \citep{SORA, Veo3, Seed_2026_Seedance2, hunyuanworld_2025_tencent, Wan2025, Genie3, Wang_2026_Matrix, Shen_2026_lyra2, Ye_2026_World}. This capability has enabled applications in robot learning \citep{Nvidia_2025_Cosmos, Kim_2025_Cosmospolicy, Ye_2026_World, veorobotics_2025}, generative gaming \citep{Wang_2026_Matrix, Valevski_2025_diffusion, Tang_2025_HunyuanGameCraft2, hunyuanworld_2025_tencent, Savva_2026_Solaris}, and world dynamics simulation \citep{Veo3, SORA, Alonso_2024_diffusionworldmodelingvisual, Genie3, Shen_2026_lyra2, Wiedemer_2025_Veo}.

However, as video generation models move toward video world models, they still lack physical consistency and controllability \citep{Kang_2025_Farvideogenerationworld, Liu_2025_PhysicalAI, Motamed_2025_PhysicalPrinciples, Bansal_2025_Videophy2, Meng_2025_World, Zhang_2025_Morpheus, Li_2025_Pisa, Zhang_2025_worldinworld}. A video world model should make future motion physically valid and controllable, not just visually realistic. Recent work has added physical priors to video generation \citep{Lv_2024_Gpt4motion, Yang_2025_VLIPP, Pandey_2025_MotionMode, Xue_2025_PhyT2V, Wang_2025_Wisa, Li_2025_Pisa, Chefer_2025_VideoJam, Yuan_2025_NewtonGen, Yuan_2025_SeeU}. These methods make motion more physically plausible. However, they do not further optimize the motion under physical constraints. As shown in Figure~\ref{fig:teaser}, consider generating a video where a cup full of water is moved from one side of a laptop to the other. There are many possible paths from the start to the goal. Some paths may reach the goal but are unsafe, not smooth, or inefficient.

In this paper, we propose \textbf{OptiWorld}, a framework that brings classical \textbf{optimal control} \citep{Stengel_1994_Optimal} into video generation at inference time. We observe that recent 3D-aware video generation models \citep{Wang_2025_Levitor, Burgert_2025_Gowiththeflow, Gu_2025_Das, Wang_2025_ATI, Chen_2025_Perception, Xing_2025_Motioncanvas, Lee_2025_Editbytrack, Burgert_2025_MotionV2V} already have strong world rendering ability, but they rely on a predefined input trajectory to specify how the world should evolve. Therefore, the key missing piece for video world models is not flawless visuals, but proactive world understanding and optimal planning before generation \citep{Zhang_2025_worldinworld}. OptiWorld fills this gap with a three-stage pipeline: understanding, planning, and generation.

The understanding stage extracts a compact, task-relevant world state from the initial frame and the goal in a zero-shot manner. It uses generic vision foundation models to obtain geometric \citep{Wang_2025_moge2} and semantic \citep{Carion_2025_SAM3} information, and a vision-language model (VLM) \citep{qwen2.5-VL} to reason about high-level object relations in the world. The planning stage is the core optimal-control component because it decides which motion the video model should render. This is challenging in open visual scenes: the planner must jointly model continuous 3D geometry and task-dependent physical constraints, including goal reaching, safety, smoothness, and efficiency. Traditional rule-based planners are often tied to fixed state spaces and hand-designed rules, while directly optimizing these heterogeneous constraints as separate penalties can be brittle and highly non-convex \citep{Wu_2025_VLAAN, Xu_2026_diffOG, Karaman_2011_RRT, Hart_1968_Astar, Rawlings_2017_MPC, Bullo_2005_Geometric}. We therefore use a Riemannian manifold as a natural way to convert world information into continuous geometry, and formulate optimal planning as a geometric shortest-path problem. For example, unsafe regions become long or uphill, while goal- and efficiency-favored directions become short or downhill; the resulting geodesic-like trajectory can then be used as video guidance. Finally, the generation module renders the video based on the planned trajectory and the initial frame.

By explicitly incorporating optimal control, OptiWorld improves the motion in generated videos, making it safer, smoother, more efficient, and more physically consistent. We demonstrate its effectiveness across multiple tasks including goal-conditioned image-to-video generation, video dynamics editing, and counterfactual generation. Our core contributions are:

\begin{itemize}
    \item We introduce optimal control into video generation, achieving safer, smoother, more efficient, and physically consistent motions via proactive world understanding and optimal planning prior to generation.

    \item We model heterogeneous physical constraints as a continuous geometric manifold, enabling optimal control for open-scene video motion planning.


\end{itemize}

\section{Related Work}
\paragraph{Video Generation Models.} 
Recent progress in video generation has been largely driven by advances in diffusion models \citep{Ho_2020_DDPM, Song_2021_Scorebased} and the scaling \citep{Kaplan_2020_Scaling} of Diffusion Transformer \citep{Peebles_2023_DiT}. Building on these developments, modern video models generate frames from images, videos, text, or actions at scale \citep{SORA, Veo3, Yang_2025_Cogvideox, Kong_2024_Hunyuanvideo, Nvidia_2025_Cosmos, Wan2025, Seed_2026_Seedance2}. These approaches achieve strong visual quality and generalization, making them promising for applications such as content generation, robot data synthesis, and interactive world simulation \citep{SORA, Veo3,  Alonso_2024_diffusionworldmodelingvisual, Genie3, Shen_2026_lyra2, Wiedemer_2025_Veo, Wang_2026_Matrix, Valevski_2025_diffusion, Tang_2025_HunyuanGameCraft2, hunyuanworld_2025_tencent, Savva_2026_Solaris, Nvidia_2025_Cosmos, Kim_2025_Cosmospolicy, Ye_2026_World, veorobotics_2025}. However, these models primarily learn appearance-level \textbf{motion} from data, without explicit modeling or control of underlying \textbf{dynamics}. As a result, generated videos may appear realistic but still violate physical principles and lack precise controllability, especially in out-of-distribution scenarios \citep{Kang_2025_Farvideogenerationworld, Liu_2025_PhysicalAI, Motamed_2025_PhysicalPrinciples, Bansal_2025_Videophy2, Meng_2025_World, Li_2025_Pisa, Chefer_2025_VideoJam, Duan_2025_Worldscore, Yuan_2026_likephys}.

\paragraph{Motion-controlled Video Generation.} 
A large body of work improves controllability by introducing additional motion-related conditions, such as camera poses, motion trajectories, optical flow, or other user-specified signals \citep{He_2024_Cameractrl, Xu_2024_Camco, Hou_2024_Camtrol, Xu_2024_Cavia, Wang_2024_MotionCtrl, Courant_2024_ET, Zhang_2025_Tora, Zhang_2025_Tora2, Geng_2024_Motionprompting, Wang_2025_Levitor, Burgert_2025_Gowiththeflow, Gu_2025_Das, Wang_2025_ATI, Chen_2025_Perception, Xing_2025_Motioncanvas, Lee_2025_Editbytrack, Burgert_2025_MotionV2V}. 
While these methods enhance controllability, the control signals are typically derived from predefined inputs, and do not fundamentally address physical consistency of the generated motion.

\paragraph{Physics-aware Video Generation.}
Recent works incorporate physical priors into video generation via three main paradigms: post-hoc simulation \citep{Lin_2024_Phys4dgen, Xie_2024_Physgaussian, Tan_2024_Physmotion, Zhang_2024_Physdreamer, Hsu_2024_Autovfx}, simulation-guided generation \citep{Yuan_2023_Physdiff, Liu_2024_Physgen, Savantaira_2024_Motioncraft, Chen_2025_Physgen3d, Xie_2025_Physanimator}, and learned priors or evaluators for improving physical plausibility \citep{Li_2024_GenerativeImageDynamics, Lv_2024_Gpt4motion, Yang_2025_VLIPP, Pandey_2025_MotionMode, Xue_2025_PhyT2V,  Wang_2025_Wisa, Yuan_2025_GenPhoto, Chefer_2025_VideoJam, Yuan_2025_NewtonGen, Yuan_2025_SeeU, Yuan_2026_WMReward}. The closest to our work is VLIPP \citep{Yang_2025_VLIPP}, which uses GPT-4o \citep{GPT4O} to generate coarse motion trajectories for guiding diffusion models. However, prior methods focus on enforcing plausibility rather than explicitly optimizing dynamics under multiple objectives such as safety, smoothness, and efficiency.

\section{Preliminaries}

A world model needs two basic components: an \emph{observation model} that abstracts information from observations, and a \emph{dynamics model} that describes how the world evolves over time. From this view, optimal control and video generation are two different forms of world modeling. Table~\ref{tab:world_model_compare} compares their observation models, dynamics models, and how OptiWorld combines them.

\begin{table*}[t]
    \centering
    \caption{\textbf{World-modeling view of OptiWorld.} A concise comparison by observation model and dynamics model.}
    \label{tab:world_model_compare}
    \footnotesize
    \setlength{\tabcolsep}{4pt}
    \renewcommand{\arraystretch}{1.22}
    \begin{tabular}{C{0.11\textwidth}C{0.09\textwidth}|L{0.24\textwidth}L{0.24\textwidth}L{0.24\textwidth}}
        \toprule
        \multicolumn{2}{c|}{\textbf{Component}} & \textbf{Optimal Control} & \textbf{Video Generation (I2V)} & \textbf{OptiWorld} \\
        \midrule
        \multirow{3}{0.11\textwidth}{\centering\textbf{Observation model}}
        & Input
        & observation $y_t$
        & frame $y_0$
        & frame + goal $(y_0, g)$ \\
        \cmidrule(lr){2-5}
        & Modeling
        & observer $\mathcal{O}_{\mathrm{oc}}$
        & encoder $E_{\mathrm{VAE}}$
        & understanding $\mathcal{E}_{\phi}$ \\
        \cmidrule(lr){2-5}
        & Output
        & state $x_t$
        & token $z_0$
        & world state $\mathcal{W}$ \\
        \midrule
        \multirow{3}{0.11\textwidth}{\centering\textbf{Dynamics model}}
        & Input
        & state/control/goal $(x_t,u_t,g)$
        & token + condition $(z_0,c)$
        & world state + goal $(\mathcal{W},g)$ \\
        \cmidrule(lr){2-5}
        & Modeling
        & dynamics function $f$ 
        & DiT $p_\theta(\cdot\mid z_0,c)$
        & planner $\mathcal{P}_{\eta}$ \\
        \cmidrule(lr){2-5}
        & Output
        & controls $u^\star_{0:T-1}$ / states $x^\star_{1:T}$
        & tokens $z_{1:T}\rightarrow\hat{y}_{1:T}$
        & optimal path $\tau^\star\rightarrow\hat{y}_{1:T}$ \\
        \bottomrule
    \end{tabular}
\end{table*}

\subsection{Optimal Control as Explicit World Modeling}

Classical optimal control is a model-based paradigm that assumes an explicit world model of the dynamical system. Its observation model turns sensor observations into a latent world state, and the dynamics model predicts how the state evolves under controls \citep{Stengel_1994_Optimal}.
Let $y_t$ denote the observation, $x_t$ the world state, and $u_t$ the control. A standard formulation is
\begin{equation}
    x_t=\mathcal{O}_{\mathrm{oc}}(y_t), \qquad
    x_{t+1}=f(x_t,u_t), \qquad
    y_t=h(x_t),
    \label{eq:state_space}
\end{equation}
where $\mathcal{O}_{\mathrm{oc}}$ estimates a planning state from an observation, $h$ maps states to observations, and $f$ is the controlled dynamics. Given a goal $g$, optimal control chooses the controls by solving
\begin{equation}
    u_{0:T-1}^{\star}
    =
    \arg\min_{u_{0:T-1}}
    \phi(x_T,g) + \sum_{t=0}^{T-1} \ell(x_t,u_t;g),
    \quad
    \text{s.t. } x_{t+1}=f(x_t,u_t),\; r_i(x_t,u_t)\leq 0.
    \label{eq:optimal_control}
\end{equation}
Here $\phi$ is a terminal cost (Mayer) term, $\ell$ is a running cost (Lagrange) term, and $r_i$ represents constraints such as collision avoidance and safety. The strength of optimal control is explicit decision-making: it selects controls or states according to clear goals and constraints. Its limitation is that it assumes the planning state and dynamics are already available.

\subsection{Video Generation as Implicit World Modeling}

Video generative models can also be considered implicit world models. Their observation model maps pixels into visual tokens and back to pixels, while their dynamics model evolves those tokens in latent space. Given an initial frame $y_0$ and a generic condition $c$, a latent video generator can be written as
\begin{equation}
    z_0=E_{\mathrm{VAE}}(y_0), \qquad
    z_{1:T}\sim p_\theta(z_{1:T}\mid z_0,c), \qquad
    \hat{y}_{1:T}=D_{\mathrm{VAE}}(z_{1:T}).
    \label{eq:video_generation}
\end{equation}
Here $E_{\mathrm{VAE}}$ and $D_{\mathrm{VAE}}$ are the visual encoder and decoder, $z_t$ denotes visual tokens, and $c$ can be a text prompt, action, or goal condition. The latent dynamics $p_\theta$ is usually implemented by a Diffusion Transformer (DiT) \citep{Peebles_2023_DiT}. This resembles state-space modeling in optimal control: $z_t$ can be interpreted as a latent state, while the DiT implicitly parameterizes transitions in this latent space.

The difference is that the dynamics are hidden inside the generator rather than exposed as a function with explicit costs and constraints. This gives video models strong scene generalization and high visual quality, but it also means they lack the decision layer of optimal control. At inference time, the model is usually asked to generate a plausible future, not to optimize which future should happen under safety, smoothness, energy, or goal constraints.

\subsection{Integrating Optimal Control into Video World Generation}

OptiWorld integrates these two views by introducing an explicit intermediate planning layer between optimal control and video generation. Its observation model converts pixels and goals into a planning-ready world state $\mathcal{W}$. Its dynamics model optimizes a trajectory $\tau$ before rendering the video. Let $J(\tau;\mathcal{W},g)$ be the physical planning objective under world state $\mathcal{W}$ and goal $g$. The decomposition is
\begin{equation}
    \mathcal{W} = \mathcal{E}_{\phi}(y_0,g), \qquad
    \tau^\star = \mathcal{P}_{\eta}(\mathcal{W},g)
    = \arg\min_{\tau \in \mathcal{C}(\mathcal{W})} J(\tau;\mathcal{W},g), \qquad
    \hat{y}_{1:T} = \mathcal{G}_{\psi}(y_0,\tau^\star,g),
    \label{eq:optiworld_decomposition}
\end{equation}
Here $\mathcal{E}_{\phi}$ is the zero-shot understanding module, and $\mathcal{W}$ contains the start state $x_0$, goal state $x_g$, and physical constraints $\mathcal{C}$. $\mathcal{P}_{\eta}$ is the optimal planner, $\tau^\star$ is the optimized trajectory, and $\mathcal{G}_{\psi}$ renders the optimized trajectory into future frames $\hat{y}_{1:T}$. In this way, OptiWorld keeps the visual generation ability of video models while adding the explicit decision layer of optimal control.

\section{Method}

\subsection{Problem Formulation}

We study physically constrained video generation for rigid-object motion in 3D scenes. The paper focuses on two tasks. The first is \emph{goal-conditioned image-to-video generation}: given an initial frame $y_0$ and a goal $g$ specified by a point and language instruction, generate a video $\hat{y}_{1:T}$ where the selected object moves along a physically preferable trajectory. The second is \emph{video dynamics editing}: given a source video, improve the object motion while preserving the scene identity and task intent. In both tasks, the generated motion should be safe, smooth, and efficient.

\subsection{Overview}

\begin{figure*}[t]
    \centering
    \includegraphics[width=\textwidth]{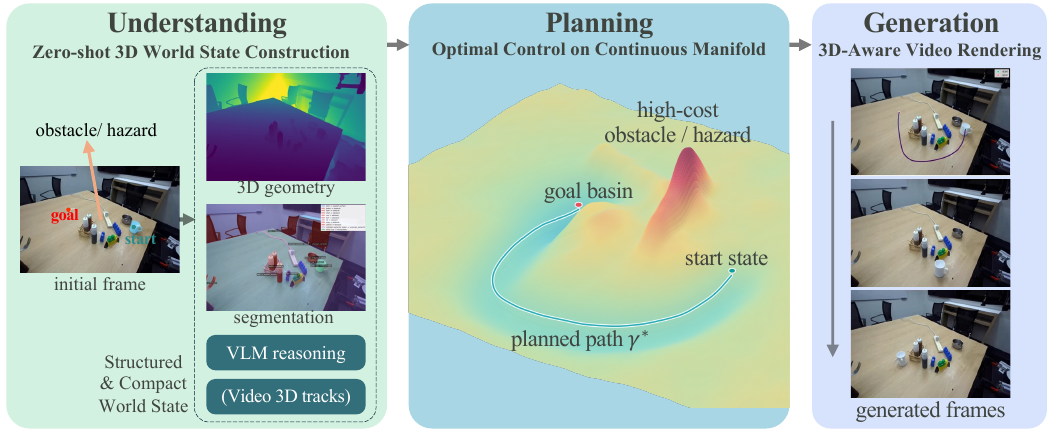}
    \caption{\textbf{OptiWorld pipeline.} Given an initial frame and goal, the understanding stage builds a compact 3D world state from geometry, segmentation, VLM reasoning, and 3D tracking (for video dynamics editing). The planning stage converts this world state into a continuous constraint-induced manifold, where hazards become high-cost regions, the goal becomes a basin, and the optimized path follows a smooth low-cost valley. The generation stage then renders detailed future frames from the first frame, prompt, and planned/optimized 3D trajectory.}
    \label{fig:pipeline}
\end{figure*}

OptiWorld has three stages, shown in Figure~\ref{fig:pipeline}. First, \emph{understanding} converts the image and goal into a task-relevant 3D world state. Second, \emph{planning} optimizes a 3D path in this world under goal-reaching, safety, smoothness, and efficiency constraints. Third, \emph{generation} uses the optimized path as motion guidance for a controllable video model.

The motivation is practical: recent 3D-trajectory-conditioned generators already have strong rendering ability once a motion signal is given \citep{Wang_2025_Levitor, Burgert_2025_Gowiththeflow, Gu_2025_Das, Wang_2025_ATI, Chen_2025_Perception, Xing_2025_Motioncanvas, Lee_2025_Editbytrack, Burgert_2025_MotionV2V}, but that motion is usually predefined by the input. OptiWorld therefore focuses on the missing steps before rendering: understanding the scene and planning a physically optimized 3D trajectory.

\subsection{Understanding: Zero-Shot 3D World State Construction}

Planning is difficult to perform directly on raw pixels when explicit geometric reasoning and constraints are required. It requires a structured scene representation that captures geometry for distance and motion reasoning, as well as semantic regions for obstacles, buffers, hazards, and targets. 
It should also encode task-level relations, since the same object can play different roles in different tasks. 
For example, a laptop is a high-risk region for a cup of water, but not necessarily for a plush toy. 
We therefore build a structured and compact world state:
\begin{equation}
    \mathcal{W} = \{\mathcal{P}, \mathcal{S}, \mathcal{A}, \mathcal{T}\}.
\end{equation}
Here $\mathcal{P}$ is the metric 3D point world, $\mathcal{S}$ contains semantic regions, $\mathcal{A}$ contains task attributes such as object role and semantic risk, and $\mathcal{T}$ contains optional 3D tracks for video editing.

We construct $\mathcal{W}$ with an agentic zero-shot pipeline. First, a metric geometry model estimates depth and 3D point maps from the input frame \citep{Wang_2025_moge2}. Second, a promptable segmentation model localizes task-relevant objects and regions \citep{Carion_2025_SAM3}. Third, a VLM reasons about the goal and assigns task roles such as obstacle, buffer, hazard, and target \citep{qwen2.5-VL}. These outputs are lifted into 3D to form object states, goal sets, signed distances, semantic maps, and task weights. For video dynamics editing, we additionally extract dense 3D point tracks \citep{Xiao_2025_Spatialtrackerv2}; these tracks become editable variables while anchoring the optimized motion to the source video. More details are provided in Appendix~\ref{app:pipeline_details}.

This design keeps the understanding module general. Each component is built from strong vision foundation models with zero-shot scene understanding ability, so OptiWorld does not need to train a new world-understanding model for every object category, scene type, or task setting. The output is not a full physical simulator, but a planning-ready state that contains just enough geometry, semantics, and task logic for the optimal planner.

\subsection{Planning: Optimal Control on a Continuous Manifold}

Planning is the core of OptiWorld because it decides which motion the video model should render. A visually strong generator cannot fix a poor trajectory: if the guidance path cuts through a hazard, jitters, or takes an unnecessary detour, the generated video will inherit those dynamics.

One possible solution is to use a traditional rule-based planning pipeline: define hand-designed costs, run a graph search such as RRT \citep{Karaman_2011_RRT}, and refine the result with a model-predictive-control (MPC)-style numerical optimizer \citep{Rawlings_2017_MPC}. This works well in specific domains where the state space and rules are fixed, such as navigation or target tracking \citep{Wu_2025_VLAAN, Xu_2026_diffOG}. However, video generation starts from open visual scenes. The planner must combine 3D geometry with task-dependent physical constraints, including goal reaching, safety, smoothness, and efficiency, all extracted or inferred from pixels. Encoding these signals as separate rules makes the modeling brittle and tied to particular scenarios, while directly optimizing them as independent penalties leads to a highly non-convex objective \citep{Bullo_2005_Geometric}.

We therefore formulate planning as a geometric problem. A Riemannian manifold gives a natural way to convert world information into continuous geometry \citep{Bullo_2005_Geometric, Ratliff_2018_RMP, Cheng_2021_RMPflow}. At each object state $q$, a positive-definite metric $G(q)$ defines the local squared travel cost
\begin{equation}
    ds_G^2 = dq^\top G(q)dq .
    \label{eq:riemannian_metric_basic}
\end{equation}
When $G(q)$ is large near unsafe regions, a path through those regions becomes long even if it is short in Euclidean distance. Therefore, planning becomes a shortest-path problem in the geometry induced by the scene, not in a fixed Euclidean space. A scalar potential $V(q)$ further turns the scene into a landscape: goals become valleys, while risky or inefficient regions become hills. This is better matched to video motion planning than flat Euclidean penalties because goal reaching, safety, efficiency, and smoothness are modeled within one scene geometry.

Formally, let $\gamma:[0,1]\rightarrow\mathcal{M}$ be a continuous path on the configuration manifold $\mathcal{M}$, with $\gamma(0)=q_s$ and $\gamma(1)\in\mathcal{G}(g)$. Its Riemannian length is
\begin{equation}
    L_G(\gamma)=\int_0^1
    \sqrt{\dot{\gamma}(s)^\top G(\gamma(s))\dot{\gamma}(s)}\,ds .
    \label{eq:riemannian_length}
\end{equation}
Minimizing the squared length energy gives a geodesic-like path under this geometry. We add the potential and dynamics terms to obtain the planning objective:
\begin{equation}
    \gamma^\star =
    \arg\min_{\gamma}
    \int_0^1
    \left[
    \frac{1}{2}\dot{\gamma}(s)^\top G(\gamma(s))\dot{\gamma}(s)
    + V(\gamma(s))
    \right]ds
    + \lambda_{\mathrm{dyn}}\int_0^1
    \left\|\nabla_{\dot{\gamma}}\dot{\gamma}(s)\right\|^2 ds,
    \label{eq:method_manifold}
\end{equation}
where $G$ is the task-induced metric, $V$ is the scalar potential, and $\nabla_{\dot{\gamma}}\dot{\gamma}$ is the covariant acceleration along the curve. Intuitively, the path is optimized as if it moves on a terrain induced by the current scene: it is pulled toward the goal, pushed away from unsafe regions, discouraged from wasting motion, and smoothed for stable video generation.

\paragraph{Constraint-induced geometry.}
All physical constraints are encoded into the same geometry. We use three planning fields: goal reaching, safety, and efficiency, plus a smoothness regularizer on the path. Let $\phi_{\mathrm{safe}}$ denote the unified safety cost from obstacles, buffers, and semantic hazards; let $\phi_{\mathrm{eff}}$ denote the efficiency cost from path length, detours, and positive lift; and let $a$ denote the gravity-up direction:
\begin{equation}
    \begin{aligned}
    G(q) =
    \bigl(1
    + \lambda_{\mathrm{safe}}\phi_{\mathrm{safe}}(q)
    + \lambda_{\mathrm{eff}}\phi_{\mathrm{eff}}(q)\bigr)I
    + \lambda_{\mathrm{up}}\phi_{\mathrm{eff}}(q)a a^\top,
    \end{aligned}
    \label{eq:metric}
\end{equation}
\begin{equation}
    \begin{aligned}
    V(q) =
    \lambda_{\mathrm{goal}}d(q,\mathcal{G}(g))^2
    + \lambda_{\mathrm{safe}}\phi_{\mathrm{safe}}(q)
    + \lambda_{\mathrm{eff}}\phi_{\mathrm{eff}}(q) .
    \end{aligned}
    \label{eq:potential}
\end{equation}
All $\lambda$ terms are nonnegative weights. The goal term creates a low-potential basin at the target. The safety term stretches or raises unsafe regions, including geometric obstacles and semantic hazards such as a laptop near a cup of water. The efficiency term favors short, low-energy motion and discourages unnecessary upward movement or detours. Smoothness is handled by the covariant acceleration term in Eq.~\eqref{eq:method_manifold}, which penalizes high acceleration, curvature, and abrupt turns. Together, these terms reshape the scene into a continuous planning manifold. Appendix~\ref{app:planning_details} gives implementation details.

\paragraph{Practical solver.}
The solver follows the geometry above in three steps. First, we convert the world state $\mathcal{W}$ into 3D cost fields for goal reaching, safety, and efficiency. Then, we search on the voxel graph to get a globally feasible seed path. Finally, we refine this seed as a continuous curve under Eq.~\eqref{eq:method_manifold}, and resample it into the frame-level trajectory $\tau^\star$ used by the generator:
\begin{equation}
    \mathcal{W}
    \rightarrow \text{3D cost fields}
    \rightarrow \text{global seed path}
    \rightarrow \gamma^\star
    \rightarrow \tau^\star .
     \label{eq:planning_path}
\end{equation}
Here $\tau^\star$ is the final optimized 3D trajectory in Euclidean coordinates.

For video dynamics editing, OptiWorld improves source 3D tracks into smoother, more efficient, and more physically reasonable motion while preserving the scene identity and task instruction. The strategy is hybrid: we combine an autonomous plan from Eq.~\eqref{eq:planning_path} with source 3D tracks to form a new global seed path, and then refine this seed. This keeps the benefits of the planned trajectory, such as efficiency and smoothness, while preserving the real motion cues already present in the original video.

\subsection{Generation: 3D-Aware Video Rendering}

Finally, the generation module takes the first frame, a text prompt, and planned frame-level tracking controls as input, and renders a photorealistic video. OptiWorld is generator-agnostic: any mature 3D-trajectory-controlled image-to-video model can use the optimized motion guidance \citep{Wang_2025_Levitor, Burgert_2025_Gowiththeflow, Gu_2025_Das, Wang_2025_ATI, Chen_2025_Perception, Xing_2025_Motioncanvas, Lee_2025_Editbytrack, Burgert_2025_MotionV2V}. In this paper, we use Diffusion as Shader (DaS) \citep{Gu_2025_Das} because it provides a convenient 3D tracking interface. For goal-conditioned image-to-video generation, the tracking controls come from the planned trajectory $\tau^\star$ of the selected object. For video dynamics editing, the input first frame is the first frame of the source video, and the tracking controls come from the optimized tracks.

\section{Experiments}
\label{sec:experiments}


\subsection{Implementation Details}
\label{sec:implementation_details}

\paragraph{Benchmark.}
We evaluate the effectiveness of OptiWorld on two video generation tasks: goal-conditioned image-to-video (I2V) generation and video dynamics editing.
For goal-conditioned I2V generation, we build a benchmark OptiBench-I2V with $60$ scene images. The scenes contain real-world indoor robot manipulation settings from DROID \citep{Khazatsky_2024_Droid} and daily-life scenes collected by us. Each example includes an initial image, a selected object, a goal point, and a language instruction. For video dynamics editing, we built OptiBench-V2V consisting of $30$ videos from DROID and our own recordings, covering robot manipulation and human hand object-moving scenarios.

\paragraph{Metrics.}
We evaluate video quality and motion quality. For video quality, we use motion smoothness, background consistency, and temporal flickering from VBench \citep{Huang_2023_Vbench}. For motion quality, we extract foreground 3D tracks from the generated videos and evaluate four aspects: goal reaching, safety, smoothness, and efficiency. For goal reaching, Goal err. measures the final 3D distance to the target in I2V, and Track dev. measures the average 3D deviation from the source motion after first-frame alignment in V2V. For safety, Viol.@succ. measures the fraction of videos that still enter unsafe regions. For smoothness, Accel. and Jerk measure second- and third-order changes of the 3D tracks. For efficiency, Path len. measures the total 3D travel distance, and Energy measures the accumulated motion cost along the track. Details are provided in the Appendix~\ref{app:metric_details}.

\paragraph{Inference details.}
Our method follows a fully inference-time pipeline. In the understanding stage, we estimate 3D geometry using MoGe2 \citep{Wang_2025_moge2} to obtain metric depth and point maps. For prompt-based segmentation, we adopt SAM3 \citep{Carion_2025_SAM3}, which provides fine-grained masks for all scene pixels and extends the vocabulary to open-world settings. For VLM reasoning, we adopt Qwen2.5-VL-3B-Instruct \citep{qwen2.5-VL}. For video dynamics editing requiring dense 3D tracks, we use SpatialTrackerV2 \citep{Xiao_2025_Spatialtrackerv2}.
The planner is the core component, solving optimal control on a geometric manifold. The understanding and planning stages take approximately $2$ minutes per scene on a single Nvidia A100-80GB GPU.
For generation, we build upon the object manipulation pipeline of Diffusion as Shader (DaS) \citep{Gu_2025_Das}, enhance its tracking video rendering for real-world 3D motion, and use $30$ denoising steps.

\subsection{Goal-Conditioned Image-to-Video Generation}
\label{sec:i2v_experiment}

Given a scene image, a goal point, and an instruction, the goal-conditioned I2V task is to move the selected object to the target while respecting physical constraints. We compare OptiWorld with state-of-the-art open-source I2V models HunyuanVideo-1.5 \citep{hunyuanvideo_2025} and Wan2.2 \citep{Wan2025}, the world foundation model Cosmos-Predict2.5 (Image2World) \citep{Nvidia_2025_CosmosPredict}, and the physics-aware video generator VLIPP \citep{Yang_2025_VLIPP}.

\begin{figure*}[t]
    \centering
    \includegraphics[width=\textwidth]{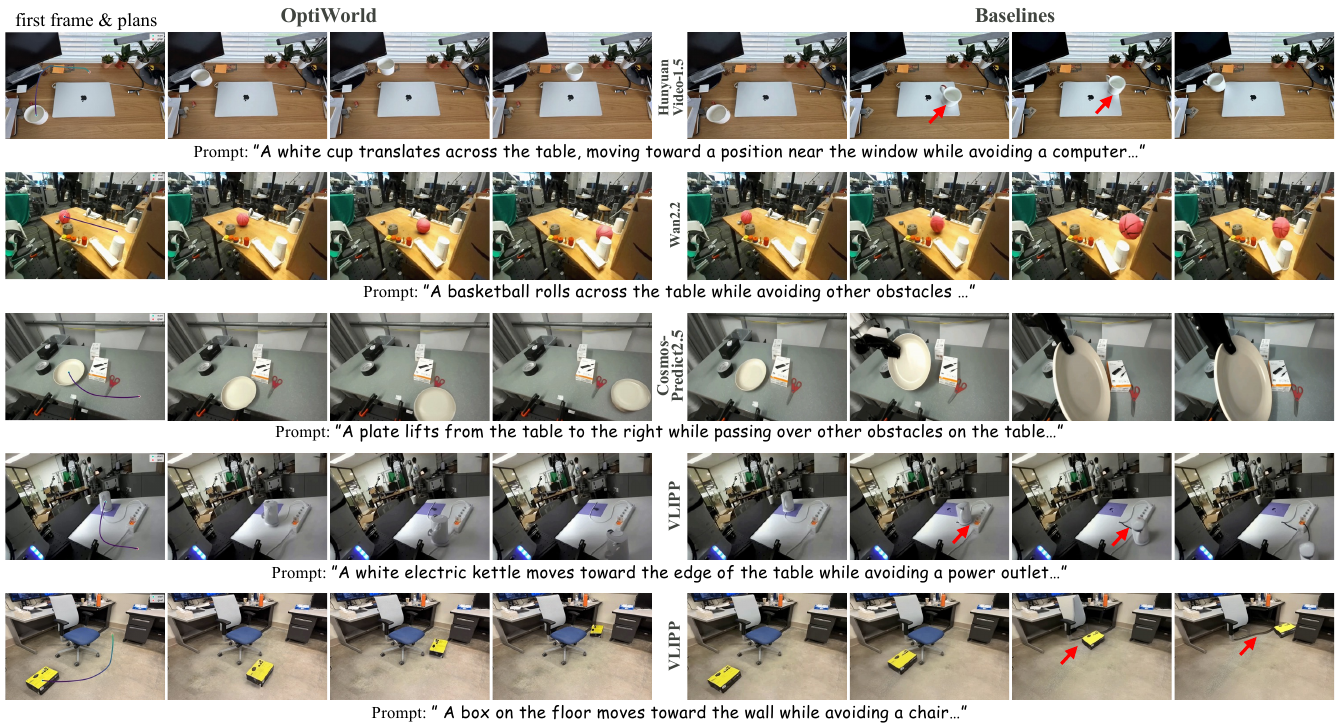}
    \caption{\textbf{Visual comparisons on goal-conditioned I2V.} OptiWorld generates videos with better goal reaching, safety, smoothness, and efficiency.}
    \label{fig:i2v_compare}
\end{figure*}

In Figure~\ref{fig:i2v_compare}, HunyuanVideo-1.5, Wan2.2, and Cosmos-Predict2.5 can generate plausible frames, but they lack explicit scene-aware planning and often fail to reach the goal accurately or safely. VLIPP introduces physics-aware guidance, but its coarse motion prior can still produce hazardous artifacts around outlet and chair. OptiWorld first plans a 3D path under these constraints and then uses it as tracking guidance, leading to more direct goal reaching, fewer safety violations, and smoother trajectories. Table~\ref{tab:i2v_results} further shows that OptiWorld achieves the best goal error, safety-violation rate, acceleration, jerk, energy, and VBench video-quality scores.

\begin{table*}[t]
    \caption{\textbf{Quantitative comparisons on goal-conditioned I2V.} Best results are highlighted in \colorbox{red!20}{red}, and second-best results in \colorbox{blue!20}{blue}.}
    \label{tab:i2v_results}
    \centering
    \footnotesize
    \setlength{\tabcolsep}{3pt}
    \setlength{\fboxsep}{1pt}
    \resizebox{\textwidth}{!}{%
    \begin{tabular}{lcccccccc}
        \toprule
        \multirow{2}{*}{Method} & \multicolumn{5}{c}{Motion quality} & \multicolumn{3}{c}{Video quality (VBench)} \\
        \cmidrule(lr){2-6}\cmidrule(lr){7-9}
        & Goal err. $\downarrow$ & Viol.@succ. $\downarrow$ & Accel. $\downarrow$ & Jerk $\downarrow$ & Energy $\downarrow$
        & Mot. smooth $\uparrow$ & BG cons. $\uparrow$ & Flicker $\uparrow$ \\
        \midrule
        HunyuanVideo-1.5 \citep{hunyuanvideo_2025} & 0.548 & 0.523 & 0.0118 & 0.0197 & 0.0576 & \colorbox{blue!20}{0.996} & 0.961 & \colorbox{blue!20}{0.995} \\
        Wan2.2 \citep{Wan2025} & 0.549 & 0.523 & 0.0172 & 0.0304 & 0.0860 & 0.993 & 0.958 & 0.990 \\
        Cosmos-Predict2.5 \citep{Nvidia_2025_CosmosPredict} & 0.733 & \colorbox{blue!20}{0.286} & 0.0104 & 0.0186 & \colorbox{blue!20}{0.0398} & 0.994 & 0.936 & 0.985 \\
        VLIPP \citep{Yang_2025_VLIPP} & \colorbox{blue!20}{0.331} & 0.511 & \colorbox{blue!20}{0.0099} & \colorbox{blue!20}{0.0176} & 0.0419 & 0.994 & \colorbox{blue!20}{0.964} & 0.992 \\
        OptiWorld & \colorbox{red!20}{0.310} & \colorbox{red!20}{0.260} & \colorbox{red!20}{0.0070} & \colorbox{red!20}{0.0116} & \colorbox{red!20}{0.0126} & \colorbox{red!20}{0.997} & \colorbox{red!20}{0.982} & \colorbox{red!20}{0.996} \\
        \bottomrule
    \end{tabular}
    }
\end{table*}

\subsection{Video Dynamics Editing}
\label{sec:v2v_experiment}
In video dynamics editing, we compare OptiWorld with the source video and Wan2.1 First-Last-Frame-to-Video (FLF2V) \citep{Wan2025}, which is a natural baseline for preserving the start and end states.

\begin{figure*}[t]
    \centering
    \includegraphics[width=\textwidth]{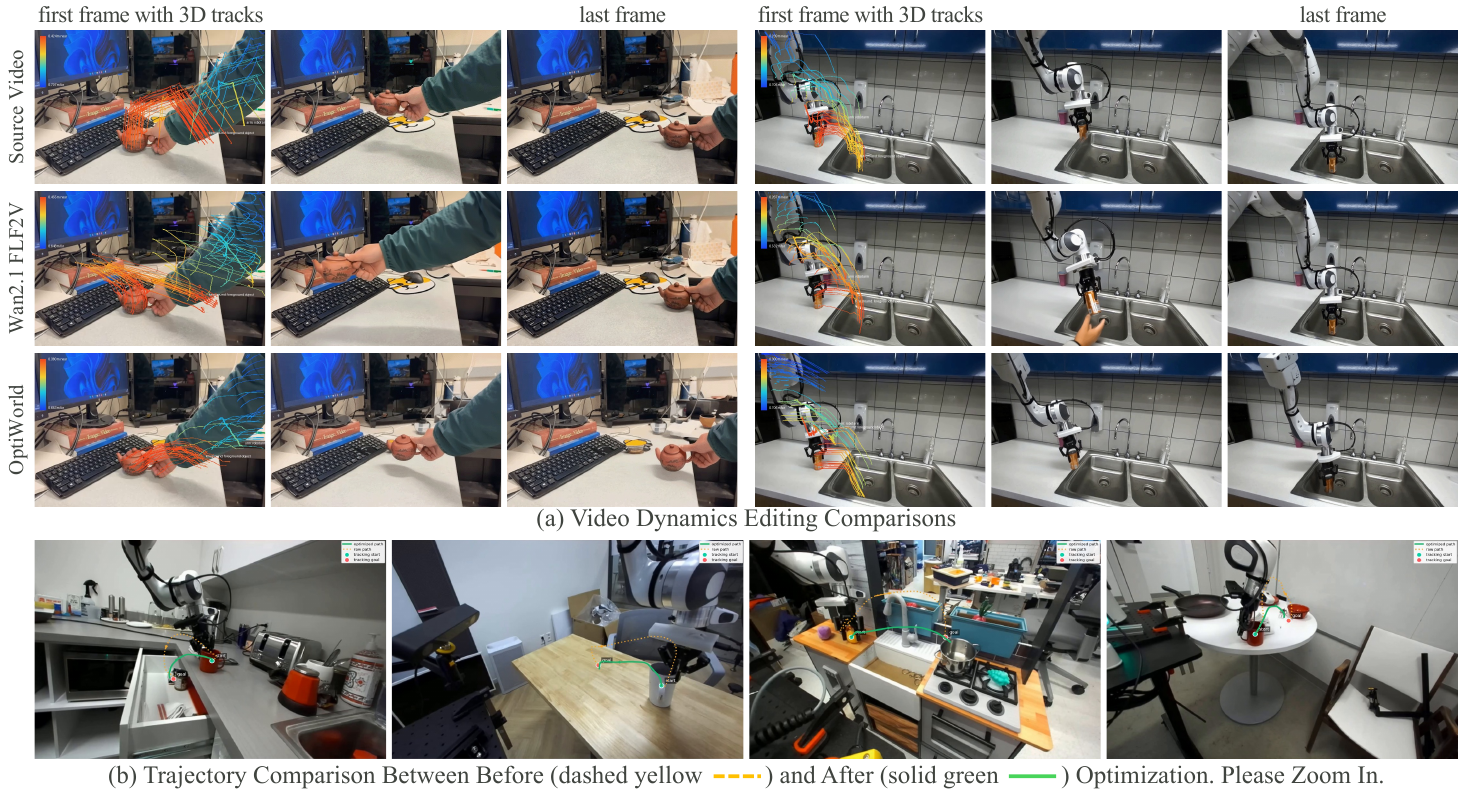}
    \caption{\textbf{Visual comparisons on video dynamics editing.} OptiWorld improves the source trajectory, producing smoother and more efficient motion.}
    \label{fig:v2v_compare}
\end{figure*}

In Figure~\ref{fig:v2v_compare}, Wan2.1 FLF2V preserves the coarse start and end frames, but the intermediate motion is not optimized and can remain long, jerky, or physically inefficient. OptiWorld instead extracts and refines the 3D tracks before generation, improving the object dynamics while preserving the scene and task intent. Table~\ref{tab:v2v_results} shows that this track-level optimization reduces track deviation, acceleration, jerk, path length, and energy without compromising video quality.

\begin{table*}[t]
    \caption{\textbf{Quantitative comparisons on video dynamics editing.}}
    \label{tab:v2v_results}
    \centering
    \footnotesize
    \setlength{\tabcolsep}{3pt}
    \setlength{\fboxsep}{1pt}
    \resizebox{\textwidth}{!}{%
    \begin{tabular}{lcccccccc}
        \toprule
        \multirow{2}{*}{Method} & \multicolumn{5}{c}{Motion quality} & \multicolumn{3}{c}{Video quality (VBench)} \\
        \cmidrule(lr){2-6}\cmidrule(lr){7-9}
        & Track dev. $\downarrow$ & Accel. $\downarrow$ & Jerk $\downarrow$ & Path len. $\downarrow$ & Energy $\downarrow$
        & Mot. smooth $\uparrow$ & BG cons. $\uparrow$ & Flicker $\uparrow$ \\
        \midrule
        Source video & -- & \colorbox{blue!20}{0.0124} & \colorbox{blue!20}{0.0205} & \colorbox{blue!20}{0.647} & \colorbox{blue!20}{0.0269} & 0.993 & \colorbox{red!20}{0.969} & 0.988 \\
        Wan2.1 FLF2V \citep{Wan2025} & \colorbox{blue!20}{0.123} & 0.0157 & 0.0266 & 0.733 & 0.0387 & \colorbox{blue!20}{0.995} & 0.957 & \colorbox{red!20}{0.993} \\
        OptiWorld & \colorbox{red!20}{0.113} & \colorbox{red!20}{0.0107} & \colorbox{red!20}{0.0181} & \colorbox{red!20}{0.524} & \colorbox{red!20}{0.0236} & \colorbox{red!20}{0.996} & \colorbox{blue!20}{0.967} & \colorbox{blue!20}{0.992} \\
        \bottomrule
    \end{tabular}
    }
\end{table*}

\subsection{Counterfactual Physics Generation}
\label{sec:counterfactual}

As shown in Figure~\ref{fig:teaser}(b), OptiWorld can also generate counterfactual motions by changing the planner before rendering. We use two simple settings. \textbf{(1) Change the goal.} For the same scene, choosing a different goal makes the planner produce a different trajectory, while the understanding and generation modules stay fixed. \textbf{(2) Change the constraints.} For the same start and goal, setting the safety constraint weight to zero can produce a dangerous path that cuts through a risky region. These two counterfactual settings can create multi-physics videos for the same scene with quantitative labels such as goal, safety violation, energy, and smoothness, which can be used to train or evaluate robot policies under controlled physical variations. More examples are provided in the Appendix~\ref{app:more_counterfactual}.

\subsection{Ablation Study}
\label{sec:ablation}

We ablate OptiWorld by pipeline stage. In understanding, w/o VLM removes active VLM reasoning and keeps only geometry and segmentation, so task-dependent object roles are no longer inferred. In planning, w/o Man. replaces the constraint-induced manifold with a flatter Euclidean planning objective; w/o Safe, w/o Eff., and w/o Smooth respectively remove the safety, efficiency, and smoothness terms from the planner. For generation, 2D guidance does not use DaS \citep{Gu_2025_Das} with our 3D trajectory guidance; instead, it projects the motion into 2D optical-flow guidance and generates videos with Go-with-the-Flow \citep{Burgert_2025_Gowiththeflow}.

Table~\ref{tab:ablations} compares each ablated variant with the full OptiWorld system on the relevant diagnostic metrics, so increases relative to OptiWorld indicate degradation. Removing VLM reasoning mainly hurts motion regularity. Removing the manifold or safety increases Viol.@succ., while removing efficiency mainly increases Energy. Removing smoothness increases Accel., Jerk, and Energy. Replacing 3D guidance with projected 2D guidance worsens the final video metrics. These results support the full pipeline design: understanding supplies task semantics, planning optimizes physical motion, and 3D generation guidance transfers the optimized motion into video.

\begin{table*}[t]
    \caption{\textbf{Quantitative comparisons on ablation study.} Planning diagnostics are computed on optimized trajectories before rendering, while video diagnostics are computed on generated videos. Dashes indicate metrics that are not the primary diagnostic for that ablation.}
    \label{tab:ablations}
    \centering
    \scriptsize
    \setlength{\tabcolsep}{3pt}
    \setlength{\fboxsep}{1pt}
    \resizebox{\textwidth}{!}{%
    \begin{tabular}{ll>{\quad}cccc>{\quad}ccc}
        \toprule
        \multirow{2}{*}{Stage} &
        \multirow{2}{*}{Variant} &
        \multicolumn{4}{c}{Planning diagnostics} &
        \multicolumn{3}{c}{Video diagnostics} \\
        \cmidrule(lr){3-6}\cmidrule(lr){7-9}
        & & Viol.@succ. $\downarrow$ & Accel. $\downarrow$ $(10^{-4})$ & Jerk $\downarrow$ $(10^{-4})$ & Energy $\downarrow$ $(10^{-4})$
        & Goal err. $\downarrow$ & Viol.@succ. $\downarrow$ & Energy $\downarrow$ \\
        \midrule
        Understanding & w/o VLM & -- & 3.10 & 1.25 & 0.35 & -- & -- & -- \\
        \midrule
        \multirow{4}{*}{Planning}
        & w/o Man. & 0.349 & -- & -- & -- & -- & -- & -- \\
        & w/o Safe & 0.589 & -- & -- & -- & -- & -- & -- \\
        & w/o Eff. & -- & -- & -- & 0.27 & -- & -- & -- \\
        & w/o Smooth & -- & 3.20 & 1.26 & 0.37 & -- & -- & -- \\
        \midrule
        Generation & 2D guidance & -- & -- & -- & -- & 0.361 & 0.523 & 0.0518 \\
        \midrule
        \multicolumn{2}{l}{OptiWorld} & 0.304 & 2.40 & 0.86 & 0.22 & 0.310 & 0.260 & 0.0126 \\
        \bottomrule
    \end{tabular}
    }
\end{table*}

\section{Conclusion}

We introduced optimal control into video generation. Through active scene understanding and manifold-based planning, OptiWorld can generate video worlds with better dynamics: motions are more controllable, smoother, safer, and more efficient. More broadly, this suggests that video world models should not only imagine plausible futures, but also choose futures that respect the structure and consequences of the physical world.

\subsubsection*{Acknowledgments}
This work is supported, in part, by the United States National Science Foundation under the grants 2133032, 2431505, and a research award from Samsung Research America. 

We thank Yichen Sheng for helpful discussion.

{
    \small
    \bibliographystyle{IEEEbib}
    \bibliography{main}

@INPROCEEDINGS{Ho_2020_DDPM,
author={Ho, Jonathan and Jain, Ajay and Abbeel, Pieter},
  booktitle={Advances in Neural Information Processing Systems}, 
  title={Denoising diffusion probabilistic models}, 
  year={2020},
  pages={6840–6851},
}

@inproceedings{Song_2021_Scorebased,
  title={Score-Based Generative Modeling through Stochastic Differential Equations},
  author={Song, Yang and Sohl-Dickstein, Jascha and Kingma, Diederik P  and Kumar, Abhishek  and Ermon, Stefano  and Poole, Ben},
  booktitle={International Conference on Learning Representations},
  year={2021},
  url={https://openreview.net/forum?id=PxTIG12RRHS}
}

@article{He_2024_Cameractrl,
      title={Camera{Ctrl}: Enabling Camera Control for Text-to-Video Generation}, 
      author={He, Hao and Xu, Yinghao and Guo, Yuwei and Wetzstein, Gordon and Dai, Bo and Li, Hongsheng and Yang, Ceyuan},
      journal={arXiv preprint arXiv:2404.02101},
      year={2024}
}

@article{Xu_2024_Camco,
  title={Cam{Co}: Camera-Controllable 3D-Consistent Image-to-Video Generation},
  author={Xu, Dejia and Nie, Weili and Liu, Chao and Liu, Sifei and Kautz, Jan and Wang, Zhangyang and Vahdat, Arash},
  journal={arXiv preprint arXiv:2406.02509},
  year={2024}
}

@article{Hou_2024_Camtrol,
  title={Training-free Camera Control for Video Generation},
  author={Hou, Chen and Wei, Guoqiang and Zeng, Yan and Chen, Zhibo},
  journal={arXiv preprint arXiv:2406.10126},
  year={2024}
}

@article{Xu_2024_Cavia,
  title={Cavia: Camera-controllable Multi-view Video Diffusion with View-Integrated Attention},
  author={Xu, Dejia and Jiang, Yifan and Huang, Chen and  Song, Liangchen and Gernoth, Thorsten and Cao, Liangliang and Wang, Zhangyang and Tang, Hao},
  journal={arXiv preprint arXiv:2410.10774},
  year={2024}
}

@inproceedings{Wang_2024_MotionCtrl,
  title={Motionctrl: A unified and flexible motion controller for video generation},
  author={Wang, Zhouxia and Yuan, Ziyang and Wang, Xintao and Li, Yaowei and Chen, Tianshui and Xia, Menghan and Luo, Ping and Shan, Ying},
  booktitle={ACM SIGGRAPH 2024 Conference Papers},
  pages={1--11},
  year={2024}
}

@article{Courant_2024_ET,
        author    = {Courant, Robin and Dufour, Nicolas and Wang, Xi and Christie, Marc and Kalogeiton, Vicky},
        title     = {{E.T. }the Exceptional Trajectories: Text-to-camera-trajectory generation with character awareness},
        journal   = {arXiv preprint arXiv:2407.01516},
        year      = {2024},
      }

@misc{GPT4O ,
  title        = {{GPT-4o}},
  author ={Openai},
  howpublished = {\url{https://openai.com/index/hello-gpt-4o/}},
}

@ARTICLE{Kaplan_2020_Scaling,
       author = {Kaplan, Jared and McCandlish,
                  Sam and Henighan,
                  Tom and Brown,
                  Tom B.  and Chess,
                  Benjamin and Child,
                  Rewon and Gray,
                  Scott and Radford,
                  Alec and Wu,
                  Jeffrey and Amodei,
                  Dario},
    title = {Scaling Laws for Neural Language Models},
    year = {2020},
    journal = {arXiv preprint arXiv: 2001.08361},
}

@inproceedings{Peebles_2023_DiT,
  title={Scalable Diffusion Models with Transformers},
  author={Peebles, William and Xie, Saining},
  booktitle = {International Conference on Computer Vision},
  year = {2023}
}

@inproceedings{Kang_2025_Farvideogenerationworld,
 author = {Kang, Bingyi and Yue, Yang and Lu, Rui and Lin, Zhijie and Zhao, Yang and Wang, Kaixin and Huang, Gao and Feng, Jiashi},
 title = {How Far is Video Generation from World Model: A Physical Law Perspective},
 booktitle = {International Conference on Machine Learning},
 year = {2025}
}

@misc{Liu_2025_PhysicalAI,
    title={Generative Physical AI in Vision: A Survey}, 
    author={Liu, Daochang and Zhang, Junyu and Dinh, Anh-Dung and Park, Eunbyung and Zhang, Shichao and Xu, Chang},
    year={2025},
    journal = {arXiv preprint arXiv: 2501.10928},
}

@misc{Motamed_2025_PhysicalPrinciples,
    title={Do generative video models understand physical principles?}, 
    author={Motamed, Saman and Culp, Laura and Swersky, Kevin and Jaini, Priyank Jaini and Geirhos, Robert},
    year={2025},
    journal = {arXiv preprint arXiv: 2501.09038},
}

@article{Zhang_2025_Morpheus,
  title={Morpheus: Benchmarking Physical Reasoning of Video Generative Models with Real Physical Experiments},
  author={Zhang, Chenyu and Cherniavskii, Daniil and Zadaianchuk, Andrii and Tragoudaras, Antonios and Vozikis, Antonios and Nijdam, Thijmen and Prinzhorn, Derck W. E. and Bodracska, Mark and Sebe, Nicu and Gavves, Efstratios},
  journal={arXiv preprint arXiv:2504.02918},
  year={2025}
}

@article{Bansal_2025_Videophy2,
  title={{VideoPhy}-2: A Challenging Action-Centric Physical Commonsense Evaluation in Video Generation},
  author={Bansal, Hritik and Peng, Clark and Bitton, Yonatan and Goldenberg, Roman and Grover, Aditya and Chang, Kai-Wei},
  journal={arXiv preprint arXiv:2503.06800},
  year={2025}
}

@article{Duan_2025_Worldscore,
    title={{WorldScore}: A Unified Evaluation Benchmark for World Generation},
    author={Duan, Haoyi and Yu, Hong-Xing and Chen, Sirui and Fei-Fei, Li and Wu, Jiajun},
    journal={arXiv preprint arXiv:2504.00983},
    year={2025}
}

@inproceedings{Meng_2025_World,
  title={Towards World Simulator: Crafting Physical Commonsense-Based Benchmark for Video Generation},
  author={Meng, Fanqing and Liao, Jiaqi and Tan, Xinyu and Shao, Wenqi and Lu, Quanfeng and Zhang, Kaipeng and Yu, Cheng and Li, Dianqi and Qiao, Yu and Luo, Ping},
  booktitle={International Conference on Machine Learning},
  year={2025},
}

@inproceedings{Li_2025_Pisa,
  title={PISA Experiments: Exploring Physics Post-Training for Video Diffusion Models by Watching Stuff Drop},
  author={Li, Chenyu and Michel, Oscar and Pan, Xichen and Liu, Sainan and Roberts, Mike and Xie, Saining},
  booktitle={International Conference on Machine Learning},
  year={2025},
}

@article{Chefer_2025_VideoJam,
  title={VideoJAM: Joint Appearance-Motion Representations for Enhanced Motion Generation in Video Models},
  author={Chefer, Hila and Singer, Uriel and Zohar, Amit and Kirstain, Yuval and Polyak, Adam and Taigman, Yaniv and Wolf, Lior and Sheynin, Shelly},
  journal={arXiv preprint arXiv:2502.02492},
  year={2025}
}

@inproceedings{Yang_2025_Cogvideox,
  title={CogVideoX: Text-to-Video Diffusion Models with An Expert Transformer},
  author={Yang, Zhuoyi and Teng, Jiayan and Zheng, Wendi and Ding, Ming and Huang, Shiyu and Xu, Jiazheng and Yang, Yuanming and Hong, Wenyi and Zhang, Xiaohan and Feng, Guanyu and others},
  booktitle={International Conference on Learning Representations},
  year={2025}
}

@article{Kong_2024_Hunyuanvideo,
  title={Hunyuanvideo: A systematic framework for large video generative models},
  author={Kong, Weijie and Tian, Qi and Zhang, Zijian and Min, Rox and Dai, Zuozhuo and Zhou, Jin and Xiong, Jiangfeng and Li, Xin and Wu, Bo and Zhang, Jianwei and others},
  journal={arXiv preprint arXiv:2412.03603},
  year={2024}
}

@article{Nvidia_2025_Cosmos,
  title={World Simulation with Video Foundation Models for Physical AI},
  author={NVIDIA},
  journal={arXiv preprint arXiv:2501.03575},
  year={2025}
}

@article{Nvidia_2025_CosmosPredict,
  title={Cosmos World Foundation Model Platform for Physical AI},
  author={NVIDIA},
  journal={arXiv preprint arXiv:2511.00062},
  year={2025}
}

@article{Lin_2024_Phys4dgen,
      title={Phys4DGen: A Physics-Driven Framework for Controllable and Efficient 4D Content Generation from a Single Image},
      author={Lin, Jiajing and Wang, Zhenzhong and Jiang, Shu and Hou, Yongjie and Jiang, Min},
      journal={arXiv preprint arXiv:2411.16800},
      year={2024}
}

@inproceedings{Xie_2024_Physgaussian,
      title={PhysGaussian: Physics-Integrated 3D Gaussians for Generative Dynamics}, 
      author={Xie, Tianyi and Zong, Zeshun and Qiu, Yuxing and Li, Xuan and Feng, Yutao and Yang, Yin and Jiang, Chenfanfu},
      booktitle = {IEEE/CVF Conference on Computer Vision and Pattern Recognition},
      year={2024},
}

@article{Tan_2024_Physmotion,
      title={PhysMotion: Physics-Grounded Dynamics From a Single Image}, 
      author={Tan, Xiyang and Jiang, Ying and Li, Xuan and Zong, Zeshun and Xie, Tianyi and Yang, Yin and Jiang, Chenfanfu},
      journal={arXiv preprint arXiv:2411.17189},
      year={2024}
}

@inproceedings{Zhang_2024_Physdreamer,
  title={{PhysDreamer}: Physics-Based Interaction with 3D Objects via Video Generation},
  author={Zhang, Tianyuan and Yu, Hong-Xing and Wu, Rundi and
         Feng, Brandon Y. and Zheng, Changxi and Snavely, Noah and Wu, Jiajun and Freeman, William T.},
  booktitle={European Conference on Computer Vision},
  year={2024}
}

@article{Hsu_2024_Autovfx,
    title={AutoVFX: Physically Realistic Video Editing from Natural Language Instructions},
    author={Hsu, Hao-Yu and Lin, Zhi-Hao and Zhai, Albert and Xia, Hongchi and Wang, Shenlong},
    journal={arXiv preprint arXiv:2411.02394},
    year={2024}
}

@inproceedings{Liu_2024_Physgen,
  title={PhysGen: Rigid-Body Physics-Grounded Image-to-Video Generation},
  author={Liu, Shaowei and Ren, Zhongzheng and Gupta, Saurabh and Wang, Shenlong},
  booktitle={European Conference on Computer Vision},
  year={2024}
}

@inproceedings{Chen_2025_Physgen3d,
  author    = {Chen, Boyuan and Jiang, Hanxiao and Liu, Shaowei and Gupta, Saurabh and Li, Yunzhu and Zhao, Hao and Wang, Shenlong},
  title     = {PhysGen3D: Crafting a Miniature Interactive World from a Single Image},
  booktitle = {IEEE/CVF Conference on Computer Vision and Pattern Recognition},
  year      = {2025},
}

@inproceedings{Xie_2025_Physanimator,
        title={PhysAnimator: Physics-Guided Generative Cartoon Animation},
        author={Xie, Tianyi and Zhao, Yiwei and Jiang, Ying and Jiang, Chenfanfu},
        booktitle = {IEEE/CVF Conference on Computer Vision and Pattern Recognition},
        year={2025}
      }

@inproceedings{Yuan_2023_Physdiff,
  title={Physdiff: Physics-guided human motion diffusion model},
  author={Yuan, Ye and Song, Jiaming and Iqbal, Umar and Vahdat, Arash and Kautz, Jan},
  booktitle = {International Conference on Computer Vision},
  year = {2023}
}

@inproceedings{Savantaira_2024_Motioncraft,
      title={MotionCraft: Physics-based Zero-Shot Video Generation},
      author={Savant Aira, Luca and Montanaro, Antonio and Aiello, Emanuele and Valsesia, Diego and Magli, Enrico},
      booktitle={Advances in Neural Information Processing Systems},
      year={2024}
}

@inproceedings{Li_2024_GenerativeImageDynamics,
      title     = {Generative Image Dynamics},
      author    = {Li, Zhengqi and Tucker, Richard and Snavely, Noah and Holynski, Aleksander},
      booktitle = {IEEE/CVF Conference on Computer Vision and Pattern Recognition},
      year      = {2024}
}

@article{Yang_2025_VLIPP,
  title={VLIPP: Towards Physically Plausible Video Generation with Vision and Language Informed Physical Prior},
  author={Yang, Xindi and Li, Baolu and Zhang, Yiming and Yin, Zhenfei and Bai, Lei and Ma, Liqian and Wang, Zhiyong and Cai, Jianfei and Wong, Tien-Tsin and Lu, Huchuan and Jia, Xu},
  journal={arXiv preprint arXiv:2503.23368},
  year={2025}
}

@inproceedings{Pandey_2025_MotionMode,
        title={Motion Modes: What Could Happen Next?},
        author={Pandey, Karran and Gadelha, Matheus and Hold-Geoffroy, Yannick and Singh, Karan and J. Mitra, Niloy and Guerrero, Paul},
        booktitle = {IEEE/CVF Conference on Computer Vision and Pattern Recognition},
        year={2025}
      }

@inproceedings{Xue_2025_PhyT2V,
        title={Phy{T2V}: LLM-Guided Iterative Self-Refinement for Physics-Grounded Text-to-Video Generation},
        author={Xue, Qiyao and Yin, Xiangyu and Yang, Boyuan and Gao, Wei},
        booktitle = {IEEE/CVF Conference on Computer Vision and Pattern Recognition},
        year={2025}
      }

@inproceedings{Lv_2024_Gpt4motion,
        title={Gpt4motion: Scripting physical motions in text-to-video generation via blender-oriented gpt planning},
        author={Lv, Jiaxi and Huang, Yi Huang and Yan, Mingfu and Huang, Jiancheng and Liu, Jianzhuang and Liu, Yifan Liu and Wen, Yafei and Chen, Xiaoxin and Chen, Shifeng},
        booktitle = {IEEE/CVF Conference on Computer Vision and Pattern Recognition},
        year={2024}
      }

@article{Yuan_2025_GenPhoto,
  title={Generative Photography: Scene-Consistent Camera Control for Realistic Text-to-Image Synthesis},
  author={Yuan, Yu and Wang, Xijun and Sheng, Yichen and Chennuri, Prateek and Zhang, Xingguang and Chan, Stanley},
  journal={IEEE/CVF Conference on Computer Vision and Pattern Recognition},
  year={2025}
}

@article{Wang_2025_Wisa,
    title={{WISA}: World Simulator Assistant for Physics-Aware Text-to-Video Generation},
    author={Wang, Jing and Ma, Ao and Cao, Ke and Zheng, Jun and Zhang, Zhanjie and Feng, Jiasong and Liu, Shanyuan and Ma, Yuhang and Cheng, Bo and Leng, Dawei and Yin, Yuhui and Liang, Xiaodan},
    journal={arXiv preprint arXiv:2502.08153},
    year={2025}
    }

@inproceedings{Zhang_2025_Tora,
        title={Tora: Trajectory-oriented Diffusion Transformer for Video Generation},
        author={Zhang, Zhenghao and Liao, Junchao and Li, Menghao and Dai, Zuozhuo and Qiu, Bingxue and Zhu, Siyu and Qin, Long and Wang, Weizhi},
        booktitle = {IEEE/CVF Conference on Computer Vision and Pattern Recognition},
        year={2025}
      }

@inproceedings{Zhang_2025_Tora2,
      title={Tora2: Motion and Appearance Customized Diffusion Transformer for Multi-Entity Video Generation}, 
      author={Zhang, Zhenghao and Liao, Junchao and Meng, Xiangyu and Qin, Long and Wang, Weizhi},
      booktitle = {ACM International Conference on Multimedia},
      year={2025}
}

@article{Geng_2024_Motionprompting,
  author    = {Geng, Daniel and Herrmann, Charles and Hur, Junhwa and Cole, Forrester and Zhang, Serena and Pfaff, Tobias and Lopez-Guevara, Tatiana and Doersch, Carl and Aytar, Yusuf and Rubinstein, Michael and Sun, Chen and Wang, Oliver and Owens, Andrew and Sun, Deqing},
  title     = {Motion Prompting: Controlling Video Generation with Motion Trajectories},
  journal   = {arXiv preprint arXiv:2412.02700},
  year      = {2024},
}

@inproceedings{Wang_2025_Levitor, 
	title={LeviTor: 3D Trajectory Oriented Image-to-Video Synthesis}, 
	author={Wang, Hanlin and Ouyang, Hao and Wang, Qiuyu and Wang, Wen and Cheng, Ka Leong and Chen, Qifeng and Shen, Yujun and Wang, Limin}, 
    booktitle = {IEEE/CVF Conference on Computer Vision and Pattern Recognition},
    year={2025}
    }

@inproceedings{Burgert_2025_Gowiththeflow,
  title={Go-with-the-Flow: Motion-Controllable Video Diffusion Models Using Real-Time Warped Noise},
  author={Burgert, Ryan and Xu, Yuancheng and Xian, Wenqi and Pilarski, Oliver and Clausen, Pascal and He, Mingming and Ma, Li and Deng, Yitong and Li, Lingxiao and Mousavi, Mohsen and Ryoo, Michael and Debevec, Paul and Yu, Ning},
  year={2025},
  booktitle={IEEE/CVF Conference on Computer Vision and Pattern Recognition},
}

@article{Gu_2025_Das,
        title={Diffusion as Shader: 3D-aware Video Diffusion for Versatile Video Generation Control}, 
         author={Gu, Zekai and Yan, Rui and Lu, Jiahao and Li, Peng and Dou, Zhiyang and Si, Chenyang and Dong, Zhen and Liu, Qifeng and Lin, Cheng and Liu, Ziwei and Wang, Wenping and Liu, Yuan},
         year={2025},
         journal={arXiv preprint arXiv:2501.03847}
        }

@article{Wang_2025_ATI,
  title={{ATI}: Any Trajectory Instruction for Controllable Video Generation},
  author={Wang, Angtian and Huang, Haibin and Fang, Zhiyuan and Yang, Yiding and Ma, Chongyang},
  journal={arXiv preprint arXiv:2505.22944},
  year={2025}
}

@article{Chen_2025_Perception,
    title={Perception-as-Control: Fine-grained Controllable Image Animation with 3D-aware Motion Representation},
    author={Chen, Yingjie and Men, Yifang and Yao, Yuan and Cui, Miaomiao and Bo, Liefeng},
    journal={arXiv preprint arXiv:2501.05020},
    year={2025}
    }

@article{Xing_2025_Motioncanvas,
  title={Motioncanvas: Cinematic shot design with controllable image-to-video generation},
  author={Xing, Jinbo and Mai, Long and Ham, Cusuh and Huang, Jiahui and Mahapatra, Aniruddha and Fu, Chi-Wing and Wong, Tien-Tsin and Liu, Feng},
  journal={arXiv preprint arXiv:2502.04299},
  year={2025}
}

@article{Lee_2025_Editbytrack,
  author    = {Lee, Yao-Chih and Zhang, Zhoutong and Huang, Jiahui and Wang, Jui-Hsien and Lee, Joon-Young and Huang, Jia-Bin and Shechtman, Eli and Li, Zhengqi},
  title     = {Generative Video Motion Editing with 3D Point Tracks},
  journal   = {arXiv preprint arXiv:2512.02015},
  year      = {2025},
}

@article{Burgert_2025_MotionV2V,
      title={MotionV2V: Editing Motion in a Video},
      author={Burgert, Ryan and Herrmann, Charles and Cole, Forrester and Ryoo, Michael S and Wadhwa, Neal and Voynov, Andrey and Ruiz, Nataniel},
      year={2025},
      journal   = {arXiv preprint arXiv:2511.20640}
}

@misc{SORA,
  title        = {Video generation models as world simulators},
   author={OpanAI},  
    howpublished = {\url{https://openai.com/index/video-generation-models-as-world-simulators/}},
}

@article{Yuan_2025_SeeU,
  title={{SeeU}: Seeing the Unseen World via 4D Dynamics-aware Generation},
  author={Yuan, Yu and Wickremasinghe, Tharindu and Nadir, Zeeshan and Wang, Xijun and Chi, Yiheng and Chan, Stanley H.},
  journal={arXiv preprint arXiv: 2512.03350},
  year={2025}
}

@article{Yuan_2025_NewtonGen,
  title={{NewtonGen}: Physics-Consistent and Controllable Text-to-Video Generation via Neural Newtonian Dynamics},
  author={Yuan, Yu and Wang, Xijun and Wickremasinghe, Tharindu and Nadir, Zeeshan and Ma, Bole and Chan, Stanley H.},
  journal={arXiv preprint arXiv: 2509.21309},
  year={2025}
}

@article{Carion_2025_SAM3,
  title={SAM 3: Segment Anything with Concepts},
  author={Carion, Nicolas and Gustafson, Laura and Hu, Yuan-Ting and Debnath, Shubham and Hu, Ronghang and Suris, D\'{i}dac and Ryali, Chaitanya and Alwala, Kalyan Vasudev and Khedr, Haitham and Huang, Anqi and Lei, Jiawen and Ma, Tengyu and Guo, Bowen and Kalla, Aditya and Marks, Matthew and Greer, John and Wang, Muxin and Sun, Peize and R{\"a}dle, Roman and Afouras, Triantafyllos and Mavroudi, Eleni and Xu, Kuan and Wu, Tsung-Han and Zhou, Yitian and Momeni, Lily and Hazra, Rishav and Ding, Sifei and Vaze, Sagar and Porcher, Florian and Li, Fan and Li, Songtao and Kamath, Aishwarya and Cheng, H. K. and Doll{\'a}r, Piotr and Ravi, Nikhila and Saenko, Kate and Zhang, Pengchuan and Feichtenhofer, Christoph},
  journal={arXiv preprint arXiv:2511.16719},
  year={2025}
}

@misc{Veo3,
  title  = {Veo 3.1: Our state-of-the-art video generation model},
   author={Google},
    year={2025},
    howpublished = {\url{https://aistudio.google.com/models/veo-3/}},
}

@article{Seed_2026_Seedance2,
  title={Seedance 2.0: Advancing Video Generation for World Complexity},
  author={Team Seedance},
  journal={arXiv preprint arXiv:2604.14148},
  year={2026}
}

@article{Wan2025,
      title={Wan: Open and Advanced Large-Scale Video Generative Models}, 
      author={Team Wan},
      journal = {arXiv preprint arXiv:2503.20314},
      year={2025}
}

@article{Wang_2025_moge2,
  title={MoGe-2: Accurate Monocular Geometry with Metric Scale and Sharp Details},
  author={Wang, Ruicheng and Xu, Sicheng and Dong, Yue and Deng, Yu and Xiang, Jianfeng and Lv, Zelong and Sun, Guangzhong and Tong, Xin and Yang, Jiaolong},
  journal={arXiv preprint arXiv:2507.02546},
  year={2025}
}

@misc{qwen2.5-VL,
    title = {Qwen2.5-VL},
    url = {https://qwenlm.github.io/blog/qwen2.5-vl/},
    author = {Qwen Team},
    year = {2025}
}

@inproceedings{Xiao_2025_Spatialtrackerv2,
  title={SpatialTrackerV2: 3D Point Tracking Made Easy},
  author={Xiao, Yuxi and Wang, Jianyuan and Xue, Nan and Karaev, Nikita and Makarov, Iurii and Kang, Bingyi and Zhu, Xin and Bao, Hujun and Shen, Yujun and Zhou, Xiaowei},
  booktitle={International Conference on Computer Vision},
  year={2025}
}

@article{Ratliff_2018_RMP,
      title={Riemannian Motion Policies}, 
      author={Ratliff, Nathan D. and Issac, Jan and Kappler, Daniel and Birchfield, Stan and Fox, Dieter},
      journal = {arXiv preprint arXiv:1801.02854},
      year={2018}
}

@ARTICLE{Cheng_2021_RMPflow,
  author={Cheng, Ching-An and Mukadam, Mustafa and Issac, Jan and Birchfield, Stan and Fox, Dieter and Boots, Byron and Ratliff, Nathan},
  journal={IEEE Transactions on Automation Science and Engineering}, 
  title={RMPflow: A Geometric Framework for Generation of Multitask Motion Policies}, 
  year={2021},
}

@book{Stengel_1994_Optimal,
  title     = {Optimal Control and Estimation},
  author    = {Stengel, Robert F.},
  publisher = {Dover Publications},
  address   = {New York, USA},
  year      = {1994}
}

@ARTICLE{Xu_2026_diffOG,
  author={Xu, Zhengtong and Miao, Zichen and Qiu, Qiang and Zhang, Zhe and She, Yu},
  journal={IEEE Transactions on Robotics}, 
  title={DiffOG: Differentiable Policy Trajectory Optimization With Generalizability}, 
  year={2026},
}

@inproceedings{Valevski_2025_diffusion,
title={Diffusion Models Are Real-Time Game Engines},
author={Valevski, Dani and Leviathan, Yaniv and Arar, Moab and Fruchter, Shlomi},
booktitle={International Conference on Learning Representations},
year={2025}
}

@article{Wang_2026_Matrix,
  title={Matrix-Game 3.0: Real-Time and Streaming Interactive World Model with Long-Horizon Memory},
  author={Wang, Zile and Liu, Zexiang and Li, Jiaxing and Huang, Kaichen and Xu, Baixin and Kang, Fei and An, Mengyin and Wang, Peiyu and Jiang, Biao and Wei, Yichen and others},
  journal={arXiv preprint arXiv:2604.08995},
  year={2026}
}

@inproceedings{Alonso_2024_diffusionworldmodelingvisual,
      title={Diffusion for World Modeling: Visual Details Matter in Atari},
      author={Alonso, Eloi and Jelley, Adam and Micheli, Vincent and Kanervisto, Anssi and Storkey, Amos and Pearce, Tim and Fleuret, François},
      booktitle={Thirty-eighth Conference on Neural Information Processing Systems},
      year={2024},
}

@article{Genie3,
  title = {Genie 3: A New Frontier for World Models},
  author = {Ball, Philip J. and Bauer, Jakob and Belletti, Frank and Brownfield, Bethanie and Ephrat, Ariel and Fruchter, Shlomi and Gupta, Agrim and Holsheimer, Kristian and Holynski, Aleksander and Hron, Jiri and Kaplanis, Christos and Limont, Marjorie and McGill, Matt and Oliveira, Yanko and Parker-Holder, Jack and Perbet, Frank and Scully, Guy and Shar, Jeremy and Spencer, Stephen and Tov, Omer and Villegas, Ruben and Wang, Emma and Yung, Jessica and Baetu, Cip and Berbel, Jordi and Bridson, David and Bruce, Jake and Buttimore, Gavin and Chakera, Sarah and Chandra, Bilva and Collins, Paul and Cullum, Alex and Damoc, Bogdan and Dasagi, Vibha and Gazeau, Maxime and Gbadamosi, Charles and Han, Woohyun and Hirst, Ed and Kachra, Ashyana and Kerley, Lucie and Kjems, Kristian and Knoepfel, Eva and Koriakin, Vika and Lo, Jessica and Lu, Cong and Mehring, Zeb and Moufarek, Alex and Nandwani, Henna and Oliveira, Valeria and Pardo, Fabio and Park, Jane and Pierson, Andrew and Poole, Ben and Ran, Helen and Salimans, Tim and Sanchez, Manuel and Saprykin, Igor and Shen, Amy and Sidhwani, Sailesh and Smith, Duncan and Stanton, Joe and Tomlinson, Hamish and Vijaykumar, Dimple and Wang, Luyu and Wingfield, Piers and Wong, Nat and Xu, Keyang and Yew, Christopher and Young, Nick and Zubov, Vadim and Eck, Douglas and Erhan, Dumitru and Kavukcuoglu, Koray and Hassabis, Demis and Ghahramani, Zoubin and Hadsell, Raia and van den Oord, A{\"a}ron and Mosseri, Inbar and Bolton, Adrian and Singh, Satinder and Rockt{\"a}schel, Tim},
  year = {2025}
}

@article{Shen_2026_lyra2,
    title={Lyra 2.0: Explorable Generative 3D Worlds},
    author={Shen, Tianchang and Bahmani, Sherwin and He, Kai and Srinivasan, Sangeetha Grama and Cao, Tianshi and Ren, Jiawei and Li, Ruilong and Wang, Zian and Sharp, Nicholas and Gojcic, Zan and Fidler, Sanja and Huang, Jiahui and Ling, Huan and Gao, Jun and Ren, Xuanchi},
    journal={arXiv preprint arXiv:2604.13036},
    year={2026}
}

@inproceedings{Kim_2025_Cosmospolicy,
  title={Cosmos Policy: Fine-Tuning Video Models for Visuomotor Control and Planning},
  author={Kim, Moo Jin and Gao, Yihuai and Lin, Tsung-Yi and Lin, Yen-Chen and Ge, Yunhao and Lam, Grace and Liang, Percy and Song, Shuran and Liu, Ming-Yu and Finn, Chelsea and Gu, Jinwei},
  booktitle={International Conference on Learning Representations (ICLR)},
  year={2026}
}

@article{veorobotics_2025,
      title={Evaluating Gemini Robotics Policies in a Veo World Simulator}, 
      author={Gemini Robotics Team},
      year={2025},
      journal={arXiv preprint arXiv:2512.10675}
}

@article{Tang_2025_HunyuanGameCraft2,
    title={Hunyuan-GameCraft-2: Instruction-following Interactive Game World Model}, 
    author={Tang, Junshu and Liu, Jiacheng and Li, Jiaqi and Wu, Longhuang and Yang, Haoyu and Zhao, Penghao and Gong, Siruis and Yuan, Xiang and Shao, Shuai and Lu, Qinglin},
    year={2025},
    journal={arXiv preprint arXiv:2511.23429},
}

@article{hunyuanworld_2025_tencent,
    title={HunyuanWorld 1.0: Generating Immersive, Explorable, and Interactive 3D Worlds from Words or Pixels},
    author={Team HunyuanWorld},
    year={2025},
    journal={arXiv preprint arXiv:2507.21809}
}

@article{Savva_2026_Solaris,
  title={Solaris: Building a Multiplayer Video World Model in Minecraft},
  author={Savva, Georgy and Michel, Oscar and Lu, Daohan and Waiwitlikhit, Suppakit and Meehan, Timothy and Mishra, Dhairya and Poddar, Srivats and Lu, Jack and Xie, Saining},
  year={2026},
  journal={arXiv preprint arXiv:2602.22208}
}

@article{Ye_2026_World,
  title={World Action Models are Zero-shot Policies},
  author={Ye, Seonghyeon and Ge, Yunhao and Zheng, Kaiyuan and Gao, Shenyuan and Yu, Sihyun and Kurian, George and Indupuru, Suneel and Tan, You Liang and Zhu, Chuning and Xiang, Jiannan and Malik, Ayaan and Lee, Kyungmin and Liang, William and Ranawaka, Nadun and Gu, Jiasheng and Xu, Yinzhen and Wang, Guanzhi and Hu, Fengyuan and Narayan, Avnish and Bjorck, Johan and Wang, Jing and Kim, Gwanghyun and Niu, Dantong and Zheng, Ruijie and Xie, Yuqi and Wu, Jimmy and Wang, Qi and Julian, Ryan and Xu, Danfei and Du, Yilun and Chebotar, Yevgen and Reed, Scott and Kautz, Jan and Zhu, Yuke and Fan, Linxi and Jang, Joel},
  journal={arXiv preprint arXiv:2602.15922},
  year={2026}
}

@article{Ha_2018_World,
  title={World Models},
  author={Ha, David and Schmidhuber, Jürgen},
  journal={arXiv preprint arXiv:1803.10122},
  year={2018}
}

@misc{Huang_2025_towards,
    title={Towards Video World Models},
    author={Huang, Xun},
    year={2025},
    url={https://www.xunhuang.me/blogs/world_model.html}
}

@article{Wiedemer_2025_Veo,
  title={Video models are zero-shot learners and reasoners},
  author={Wiedemer, Thadd{\"a}us and Li, Yuxuan and Vicol, Paul and Gu, Shixiang Shane and Matarese, Nick and Swersky, Kevin and Kim, Been and Jaini, Priyank and Geirhos, Robert},
  journal={arXiv preprint arXiv:2509.20328},
  year={2025}
}

@ARTICLE{Hart_1968_Astar,
  author={Hart, Peter E. and Nilsson, Nils J. and Raphael, Bertram},
  journal={IEEE Transactions on Systems Science and Cybernetics}, 
  title={A Formal Basis for the Heuristic Determination of Minimum Cost Paths}, 
  year={1968},
}

@book{Rawlings_2017_MPC, 
author = {Rawlings, J. and Mayne, D.Q. and Diehl, Moritz},
year = {2017},
title = {Model Predictive Control: Theory, Computation, and Design}
}

@book{Bullo_2005_Geometric, 
author = {Bullo, Francesco and Lewis, Andrew D},
year = {2005},
title = {Geometric control of mechanical systems: modeling, analysis, and design for simple mechanical control systems}
}

@article{Karaman_2011_RRT,
  title={Sampling-Based Algorithms for Optimal Motion Planning},
  author={Karaman, Sertac and Frazzoli, Emilio},
  journal={The International Journal of Robotics Research (IJRR)},
  year={2011}
}

@article{Wu_2025_VLAAN,
  title={{VLA-AN}: An Efficient and Onboard Vision-Language-Action Framework for Aerial Navigation in Complex Environments},
  author={Wu, Yuze and Zhu, Mo and Li, Xingxing and Du, Yuheng and Fan, Yuxin and Li, Wenjun and Han, Zhichao and Zhou, Xin and Gao, Fei},
  journal={arXiv preprint arXiv:2512.15258},
  year={2025}
}

@article{Zhang_2025_worldinworld,
  title        = {World-in-World: World Models in a Closed-Loop World},
  author       = {Zhang, Jiahan and Jiang, Muqing and Dai, Nanru and Lu, Taiming and Uzunoglu, Arda and Zhang, Shunchi and Wei, Yana and Wang, Jiahao and Patel, Vishal M. and Liang, Paul Pu and Khashabi, Daniel and Peng, Cheng and Chellappa, Rama and Shu, Tianmin and Yuille, Alan and Du, Yilun and Chen, Jieneng},
  year         = {2025},
  journal={arXiv preprint arXiv:2510.18135},
}

@article{Khazatsky_2024_Droid,
    title   = {DROID: A Large-Scale In-The-Wild Robot Manipulation Dataset},
    author  = {Khazatsky, Alexander and Pertsch, Karl and Nair, Suraj and Balakrishna, Ashwin and Dasari, Sudeep and Karamcheti, Siddharth and Nasiriany, Soroush and Srirama, Mohan Kumar and Chen, Lawrence Yunliang and Ellis, Kirsty and Fagan, Peter David and Hejna, Joey and Itkina, Masha and Lepert, Marion and Ma, Yecheng Jason and Miller, Patrick Tree and Wu, Jimmy and Belkhale, Suneel and Dass, Shivin and Ha, Huy and Jain, Arhan and Lee, Abraham and Lee, Youngwoon and Memmel, Marius and Park, Sungjae and Radosavovic, Ilija and Wang, Kaiyuan and Zhan, Albert and Black, Kevin and Chi, Cheng and Hatch, Kyle Beltran and Lin, Shan and Lu, Jingpei and Mercat, Jean and Rehman, Abdul and Sanketi, Pannag R and Sharma, Archit and Simpson, Cody and Vuong, Quan and Walke, Homer Rich and Wulfe, Blake and Xiao, Ted and Yang, Jonathan Heewon and Yavary, Arefeh and Zhao, Tony Z and Agia, Christopher and Baijal, Rohan and Castro, Mateo Guaman and Chen, Daphne and Chen, Qiuyu and Chung, Trinity and Drake, Jaimyn and Foster, Ethan Paul and Gao, Jensen and Guizilini, Vitor and Herrera, David Antonio and Heo, Minho and Hsu, Kyle and Hu, Jiaheng and Irshad, Muhammad Zubair and Jackson, Donovon and Le, Charlotte and Li, Yunshuang and Lin, Kevin and Lin, Roy and Ma, Zehan and Maddukuri, Abhiram and Mirchandani, Suvir and Morton, Daniel and Nguyen, Tony and O'Neill, Abigail and Scalise, Rosario and Seale, Derick and Son, Victor and Tian, Stephen and Tran, Emi and Wang, Andrew E and Wu, Yilin and Xie, Annie and Yang, Jingyun and Yin, Patrick and Zhang, Yunchu and Bastani, Osbert and Berseth, Glen and Bohg, Jeannette and Goldberg, Ken and Gupta, Abhinav and Gupta, Abhishek and Jayaraman, Dinesh and Lim, Joseph J and Malik, Jitendra and Martín-Martín, Roberto and Ramamoorthy, Subramanian and Sadigh, Dorsa and Song, Shuran and Wu, Jiajun and Yip, Michael C and Zhu, Yuke and Kollar, Thomas and Levine, Sergey and Finn, Chelsea},
    year    = {2024},
    journal = {arXiv preprint arXiv:2403.12945},
}

@article{hunyuanvideo_2025,
      title={HunyuanVideo 1.5 Technical Report}, 
      author={Tencent Hunyuan Foundation Model Team},
      year={2025},
      journal = {arXiv preprint arXiv:2511.18870},
}

@InProceedings{Huang_2023_Vbench,
     title={{VBench}: Comprehensive Benchmark Suite for Video Generative Models},
     author={Huang, Ziqi and He, Yinan and Yu, Jiashuo and Zhang, Fan and Si, Chenyang and Jiang, Yuming and Zhang, Yuanhan and Wu, Tianxing and Jin, Qingyang and Chanpaisit, Nattapol and Wang, Yaohui and Chen, Xinyuan and Wang, Limin and Lin, Dahua and Qiao, Yu and Liu, Ziwei},
     booktitle={IEEE/CVF Conference on Computer Vision and Pattern Recognition},
     year={2024}
 }

@inproceedings{Yuan_2026_likephys,
title={LikePhys: Evaluating Intuitive Physics Understanding in Video Diffusion Models via Likelihood Preference},
author={ Yuan, Jianhao and Pizzati, Fabio and Pinto, Francesco and Kunze, Lars and Laptev, Ivan and Newman, Paul and Torr, Philip and Martini, Daniele De},
booktitle={International Conference on Learning Representations},
year={2026},
}

@InProceedings{Yuan_2026_WMReward,
    title={Inference-time Physics Alignment of Video Generative Models with Latent World Models},
    author={Yuan, Jianhao and Zhang, Xiaofeng and Friedrich, Felix and Beltran-Velez, Nicolas and Hall, Melissa and Askari-Hemmat, Reyhane and Han, Xiaochuang and Ballas, Nicolas and Drozdzal, Michal and Romero-Soriano, Adriana},
    booktitle={IEEE/CVF Conference on Computer Vision and Pattern Recognition},
    year={2026}
}
}

\newpage


\appendix

\section{Benchmark Details}
\label{app:benchmark}

Table~\ref{tab:benchmark_sources} summarizes the data used in our benchmark. DROID scenes provide real robot manipulation layouts. Our collected data add daily indoor scenes and hand-object motion examples that contain semantic hazards such as laptops, outlets, chairs, fragile objects, and cluttered tables.

\begin{table}[h]
    \centering
    \caption{\textbf{Benchmark data sources.}}
    \label{tab:benchmark_sources}
    \begin{tabular}{lccc}
        \toprule
        Task & DROID & Collected by us & Total \\
        \midrule
        Goal-conditioned image-to-video generation & 22 & 38 & 60 \\
        Video dynamics editing & 20 & 10 & 30 \\
        \bottomrule
    \end{tabular}
\end{table}

\paragraph{Annotation.}
For goal-conditioned image-to-video generation, each example is annotated with a selected movable object, a goal point, and a language instruction. The selected-object mask is extracted by SAM3 \citep{Carion_2025_SAM3}. The goal point is given by the user's click.

\begin{figure*}[t]
    \centering
    \includegraphics[width=\textwidth]{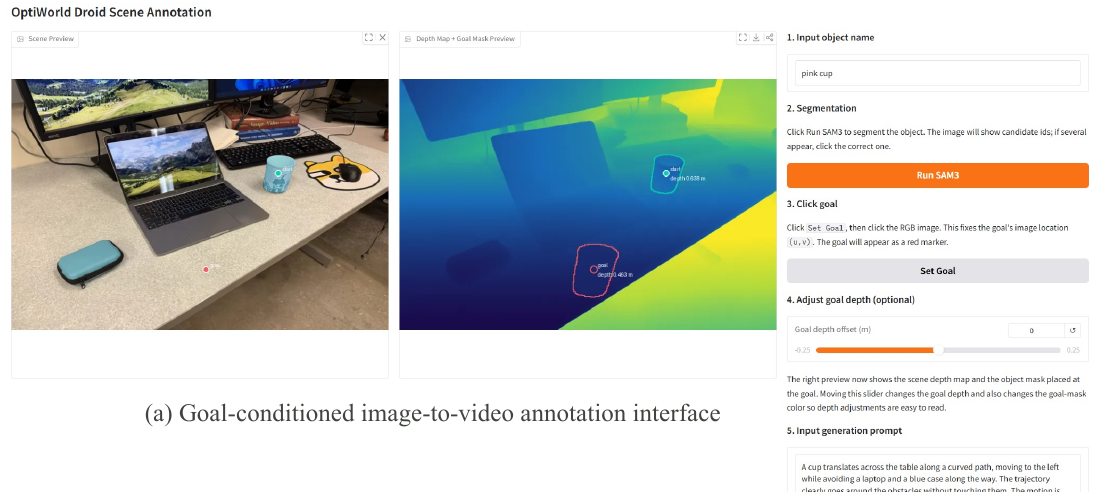}
    \caption{\textbf{Data annotation interface.}}
    \label{fig:annotation_interface}
\end{figure*}

\section{OptiWorld Pipeline Details}
\label{app:pipeline_details}

\subsection{Understanding}
\label{app:understanding_details}

The understanding stage turns an image or a video into a compact 3D semantic world state. 

\paragraph{Understanding for goal-conditioned image-to-video generation.}
For each input image, we first resize it to a fixed working resolution. MoGe2 \citep{Wang_2025_moge2} estimates metric depth, a dense 3D point map, and camera intrinsics. The annotation gives the selected object and the goal point. SAM3 \citep{Carion_2025_SAM3} refines the selected object mask and the task-relevant scene masks. Qwen2.5-VL \citep{qwen2.5-VL} reads the image and instruction, then assigns semantic roles such as manipulable object, support surface, obstacle, safety buffer, safety hazard, and target. We lift every valid mask pixel to 3D using the point map. The output contains the object centroid, goal position, semantic point cloud, signed-distance fields, hazard fields, and task weights.

\paragraph{Video dynamics editing.}
Video dynamics editing uses the same first-frame geometry and semantic reasoning, but adds source motion. We run SpatialTrackerV2 \citep{Xiao_2025_Spatialtrackerv2} on the source video to obtain dense 2D and 3D tracks, depth, point maps, intrinsics, and camera information. The foreground and arm masks select the tracks that can be edited. Background tracks are kept as anchors. The output world state therefore contains both the static scene constraints and the raw source 3D motion that should be preserved in intent but improved in dynamics.

\subsection{Planning}
\label{app:planning_details}

The planner optimizes object motion in the 3D constraint-induced manifold defined in the main text. Let $q_k\in\mathbb{R}^3$ be the object state at planning step $k$, $\mathcal{G}(g)$ be the 3D goal set, $d_{\mathrm{obs}}(q)$ be the signed distance to geometric obstacles, $d_{\mathrm{haz}}(q)$ be the distance to semantic hazards, and $a$ be the gravity-up direction.

\paragraph{Constraint fields.}
We use the same goal, safety, and efficiency fields as Eq.~\eqref{eq:metric} and Eq.~\eqref{eq:potential}. The safety field combines geometric obstacles and semantic hazards:
\begin{align}
    \phi_{\mathrm{safe}}(q)
    &= \phi_{\mathrm{obs}}(q)+\phi_{\mathrm{haz}}(q), \\
    \phi_{\mathrm{obs}}(q)
    &= \exp\!\left(-\frac{d_{\mathrm{obs}}(q)-m_{\mathrm{obs}}}{\sigma_{\mathrm{obs}}}\right), \\
    \phi_{\mathrm{haz}}(q)
    &= r(q)\exp\!\left(-\frac{d_{\mathrm{haz}}(q)-m_{\mathrm{haz}}}{\sigma_{\mathrm{haz}}}\right).
\end{align}
Here $m_{\mathrm{obs}}$ and $m_{\mathrm{haz}}$ are safety margins, $\sigma_{\mathrm{obs}}$ and $\sigma_{\mathrm{haz}}$ control the spatial falloff, and $r(q)$ is the VLM-estimated semantic risk weight. The efficiency field is
\begin{equation}
    \phi_{\mathrm{eff}}(q)=\max\bigl((q-q_s)^\top a,0\bigr)^2,
\end{equation}
which discourages unnecessary positive lift from the start state $q_s$. Path length and detours are handled by the metric length term during graph search and continuous refinement.

These fields define the local metric
\begin{equation}
    G(q)=
    \bigl(1
    + \lambda_{\mathrm{safe}}\phi_{\mathrm{safe}}(q)
    + \lambda_{\mathrm{eff}}\phi_{\mathrm{eff}}(q)\bigr)I
    + \lambda_{\mathrm{up}}\phi_{\mathrm{eff}}(q)aa^\top ,
    \label{eq:appendix_metric}
\end{equation}
and the scalar potential
\begin{equation}
    V(q)=
    \lambda_{\mathrm{goal}}d(q,\mathcal{G}(g))^2
    + \lambda_{\mathrm{safe}}\phi_{\mathrm{safe}}(q)
    + \lambda_{\mathrm{eff}}\phi_{\mathrm{eff}}(q).
    \label{eq:appendix_potential}
\end{equation}
Large safety or efficiency costs stretch the metric and raise the potential. A path through a hazard or obstacle therefore becomes expensive even if it is short in Euclidean distance, while the goal term keeps the endpoint close to $\mathcal{G}(g)$.

\paragraph{Discrete seed.}
We voxelize the 3D world and run A* \citep{Hart_1968_Astar} graph search to find a globally connected seed path. For neighboring voxels $q_i$ and $q_j$, with midpoint $q_{ij}$ and $\Delta q=q_j-q_i$, the edge cost is
\begin{equation}
    c_{ij}=
    \sqrt{\Delta q^\top G(q_{ij})\Delta q}
    +\|\Delta q\|_2 V(q_{ij}).
    \label{eq:appendix_edge_cost}
\end{equation}
This seed already avoids most unsafe shortcuts.

\paragraph{Continuous refinement.}
We resample the seed into a curve and optimize a small set of control points with field, smoothness, and anchor terms:
\begin{equation}
    E(\tau)=
    E_G
    +E_V
    +\lambda_{\mathrm{dyn}}E_{\mathrm{dyn}}
    +\lambda_{\mathrm{anchor}}E_{\mathrm{anchor}} .
    \label{eq:appendix_planning_energy}
\end{equation}
Here $E_G$ samples the Riemannian metric in Eq.~\eqref{eq:appendix_metric}, and $E_V$ samples the potential in Eq.~\eqref{eq:appendix_potential}. Thus goal reaching, safety, and efficiency keep the same symbols and weights as the main objective. The dynamics term penalizes acceleration and sharp turns:
\begin{equation}
    E_{\mathrm{dyn}}=\sum_k \|q_{k+1}-2q_k+q_{k-1}\|_2^2 .
\end{equation}
For goal-conditioned image-to-video generation, the anchor term keeps the refined path near the graph seed. For video dynamics editing, it anchors the refined motion to the source tracks so that the task intent is preserved. We use a gradient-based solver on the control points and then resample the optimized curve into frame-level controls for generation.

\subsection{Generation}
\label{app:generation_details}

We use Diffusion as Shader (DaS) \citep{Gu_2025_Das} as the tracking-conditioned renderer. The original DaS interface expects a tracking video as condition input. We modify the bridge so that this control video is generated from metric 3D points rather than only 2D image displacements.

For goal-conditioned image-to-video generation, we load the MoGe2 \citep{Wang_2025_moge2} point map, camera intrinsics, selected object mask, and planned 3D path. Points inside the selected object mask receive the planned rigid translation. Support-surface and background points remain static. We then project all 3D points back to pixels in each frame and render a dense tracking video. This makes the control signal camera-aware and depth-aware.

For video dynamics editing, foreground and arm tracks are initialized from SpatialTrackerV2. The optimized 3D tracks replace the raw foreground motion. Background tracks keep their source motion or remain static, depending on the camera motion. We render the optimized tracks into the same DaS tracking-video format. These changes let DaS follow the planned 3D dynamics while preserving scene appearance.

\section{Metric Details}
\label{app:metric_details}

We compute motion metrics from foreground 3D tracks. Let $p_t$ be the representative object track at frame $t$, $g$ be the 3D goal, and $S_{\mathrm{unsafe}}$ be the unsafe region.

\paragraph{Goal reaching.}
For goal-conditioned image-to-video generation, goal error is the final distance to the target:
\begin{equation}
    \mathrm{GoalError}=\|p_T-g\|_2 .
\end{equation}
For video dynamics editing, track deviation compares the generated track $\hat{p}_t$ with the source track $p_t$ after first-frame alignment:
\begin{equation}
    \mathrm{TrackDeviation}=\frac{1}{T}\sum_{t=1}^{T}\|(\hat{p}_t-\hat{p}_1)-(p_t-p_1)\|_2 .
\end{equation}

\paragraph{Safety.}
Violation among successful cases is measured only on goal-conditioned image-to-video generations that reach the goal. It is the fraction of these generations whose foreground track enters the unsafe set:
\begin{equation}
    \mathrm{ViolationAmongSuccess}=
    \frac{\sum_i \mathbb{1}[\mathrm{success}_i]\mathbb{1}[\exists t,\ p_{i,t}\in S_{\mathrm{unsafe}}]}
    {\sum_i \mathbb{1}[\mathrm{success}_i]} .
\end{equation}

\paragraph{Smoothness.}
Acceleration and jerk measure second- and third-order temporal changes:
\begin{align}
    \mathrm{Acceleration} &= \frac{1}{T-2}\sum_{t=2}^{T-1}\|p_{t+1}-2p_t+p_{t-1}\|_2, \\
    \mathrm{Jerk} &= \frac{1}{T-3}\sum_{t=2}^{T-2}\|p_{t+2}-3p_{t+1}+3p_t-p_{t-1}\|_2 .
\end{align}

\paragraph{Efficiency.}
Path length is the total traveled distance:
\begin{equation}
    \mathrm{PathLength}=\sum_{t=1}^{T-1}\|p_{t+1}-p_t\|_2 .
\end{equation}
Energy is the accumulated motion cost:
\begin{equation}
    \mathrm{Energy}=\sum_{t=1}^{T-1}\|p_{t+1}-p_t\|_2^2 .
\end{equation}
For video quality, we report VBench motion smoothness, background consistency, and temporal flickering using the official VBench implementation.

\section{More Results for Goal-Conditioned Image-to-Video Generation}
\label{app:more_goal_conditioned_image_to_video}

\begin{figure}[h]
    \centering
    \includegraphics[width=\linewidth]{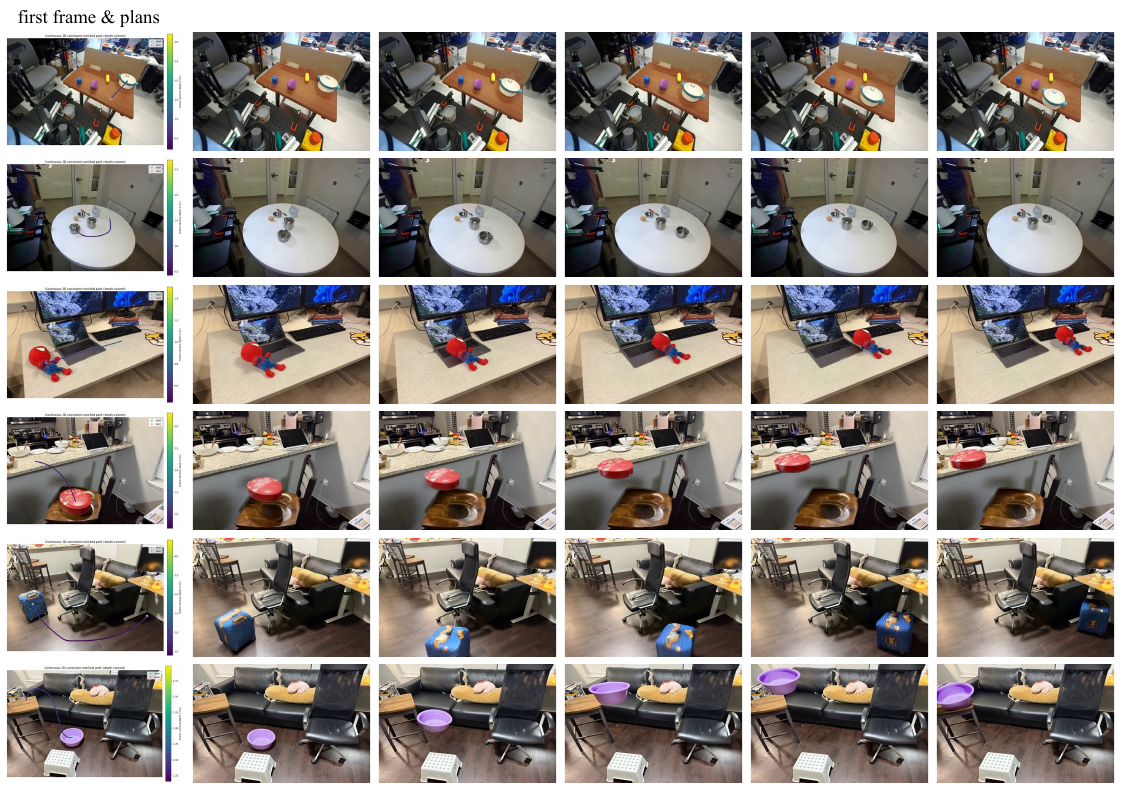}
    \caption{\textbf{Additional goal-conditioned image-to-video results.}}
    \label{fig:more_goal_conditioned_image_to_video_results}
\end{figure}

Table~\ref{tab:more_goal_conditioned_image_to_video_std} reports motion-quality results with standard deviations. Figure~\ref{fig:more_goal_conditioned_image_to_video_results} provides additional visual examples for goal-conditioned image-to-video generation.

\begin{table*}[h]
    \centering
    \caption{\textbf{Additional quantitative results on goal-conditioned image-to-video generation.} Values are mean $\pm$ standard deviation.}
    \label{tab:more_goal_conditioned_image_to_video_std}
    \footnotesize
    \setlength{\tabcolsep}{3pt}
    \resizebox{\textwidth}{!}{%
    \begin{tabular}{lccccc}
        \toprule
        Method & Goal error $\downarrow$ & Violation among successful cases $\downarrow$ & Acceleration $\downarrow$ & Jerk $\downarrow$ & Energy $\downarrow$ \\
        \midrule
        HunyuanVideo-1.5 & 0.548 $\pm$ 0.465 & 0.5232 $\pm$ 0.3114 & 0.0118 $\pm$ 0.0168 & 0.0197 $\pm$ 0.0301 & 0.0576 $\pm$ 0.1550 \\
        Wan2.2 & 0.549 $\pm$ 0.371 & 0.5230 $\pm$ 0.2607 & 0.0172 $\pm$ 0.0235 & 0.0304 $\pm$ 0.0437 & 0.0860 $\pm$ 0.2172 \\
        Cosmos-Predict2.5 & 0.733 $\pm$ 0.375 & 0.2857 $\pm$ 0.2440 & 0.0104 $\pm$ 0.0134 & 0.0186 $\pm$ 0.0256 & 0.0398 $\pm$ 0.1409 \\
        VLIPP & 0.331 $\pm$ 0.316 & 0.5110 $\pm$ 0.2694 & 0.0099 $\pm$ 0.0190 & 0.0176 $\pm$ 0.0366 & 0.0419 $\pm$ 0.2171 \\
        OptiWorld & 0.310 $\pm$ 0.317 & 0.2599 $\pm$ 0.2641 & 0.0070 $\pm$ 0.0031 & 0.0116 $\pm$ 0.0053 & 0.0126 $\pm$ 0.0100 \\
        \bottomrule
    \end{tabular}
    }
\end{table*}

\section{More Results for Video Dynamics Editing}
\label{app:more_video_dynamics_editing}

\begin{figure}[h]
    \centering
    \includegraphics[width=\linewidth]{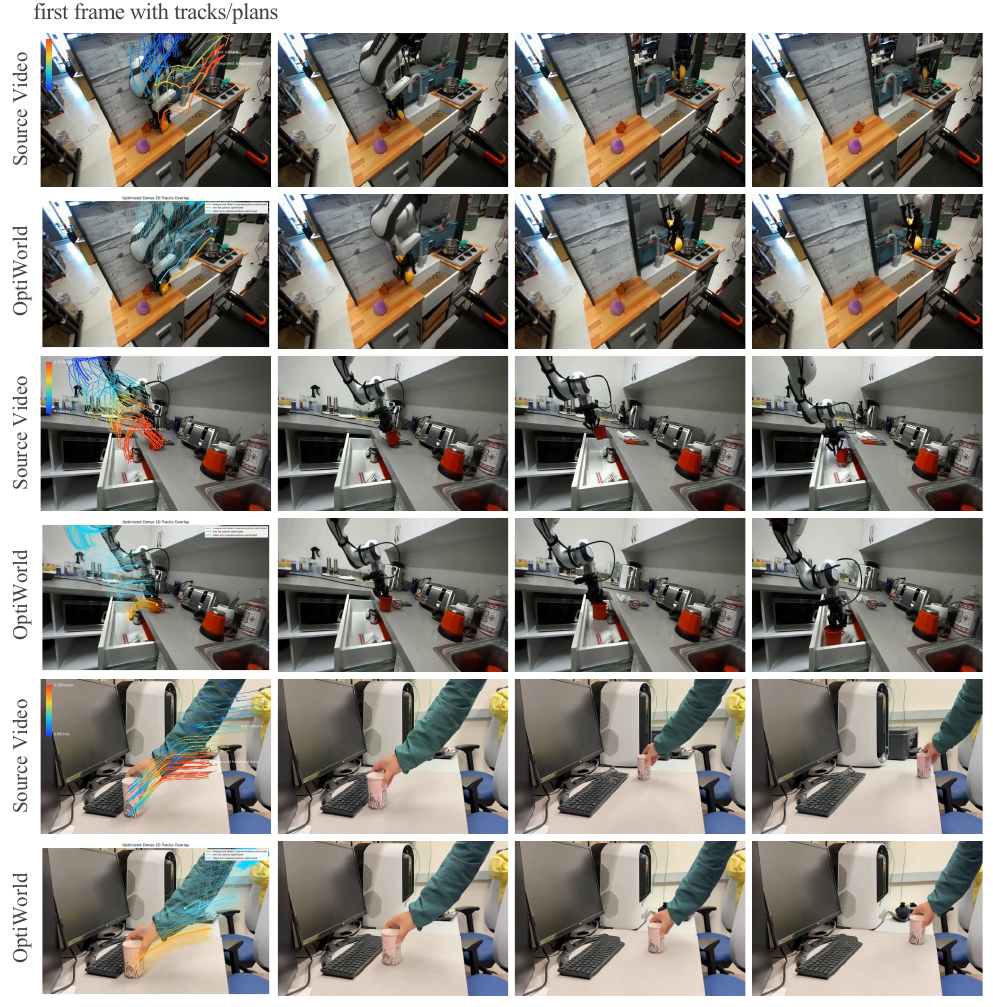}
    \caption{\textbf{Additional video dynamics editing results.} OptiWorld refines the source 3D motion before rendering, producing smoother and shorter object trajectories while keeping the first-frame scene content stable. Note that OptiWorld only relies on the \textbf{first frame}, prompt, and the optimized tracks for video dynamics editing.}
    \label{fig:more_video_dynamics_editing_results}
\end{figure}

Table~\ref{tab:more_video_dynamics_editing_std} reports video dynamics editing results with standard deviations. Figure~\ref{fig:more_video_dynamics_editing_results} provides additional visual examples for video dynamics editing.

\begin{table*}[h]
    \centering
    \caption{\textbf{Additional quantitative results on video dynamics editing.} Values are mean $\pm$ standard deviation.}
    \label{tab:more_video_dynamics_editing_std}
    \footnotesize
    \setlength{\tabcolsep}{3pt}
    \resizebox{\textwidth}{!}{%
    \begin{tabular}{lccccc}
        \toprule
        Method & Track deviation $\downarrow$ & Acceleration $\downarrow$ & Jerk $\downarrow$ & Path length $\downarrow$ & Energy $\downarrow$ \\
        \midrule
        Source video & -- & 0.0124 $\pm$ 0.0128 & 0.0205 $\pm$ 0.0229 & 0.647 $\pm$ 0.342 & 0.0269 $\pm$ 0.0398 \\
        Wan2.1 first-last-frame-to-video & 0.123 $\pm$ 0.070 & 0.0157 $\pm$ 0.0125 & 0.0266 $\pm$ 0.0235 & 0.733 $\pm$ 0.343 & 0.0387 $\pm$ 0.0431 \\
        OptiWorld & 0.113 $\pm$ 0.072 & 0.0107 $\pm$ 0.0121 & 0.0181 $\pm$ 0.0213 & 0.524 $\pm$ 0.346 & 0.0236 $\pm$ 0.0489 \\
        \bottomrule
    \end{tabular}
    }
\end{table*}

\section{More Results for Ablation Study}
\label{app:ablation_details}

\begin{figure}[h]
    \centering
    \includegraphics[width=\linewidth]{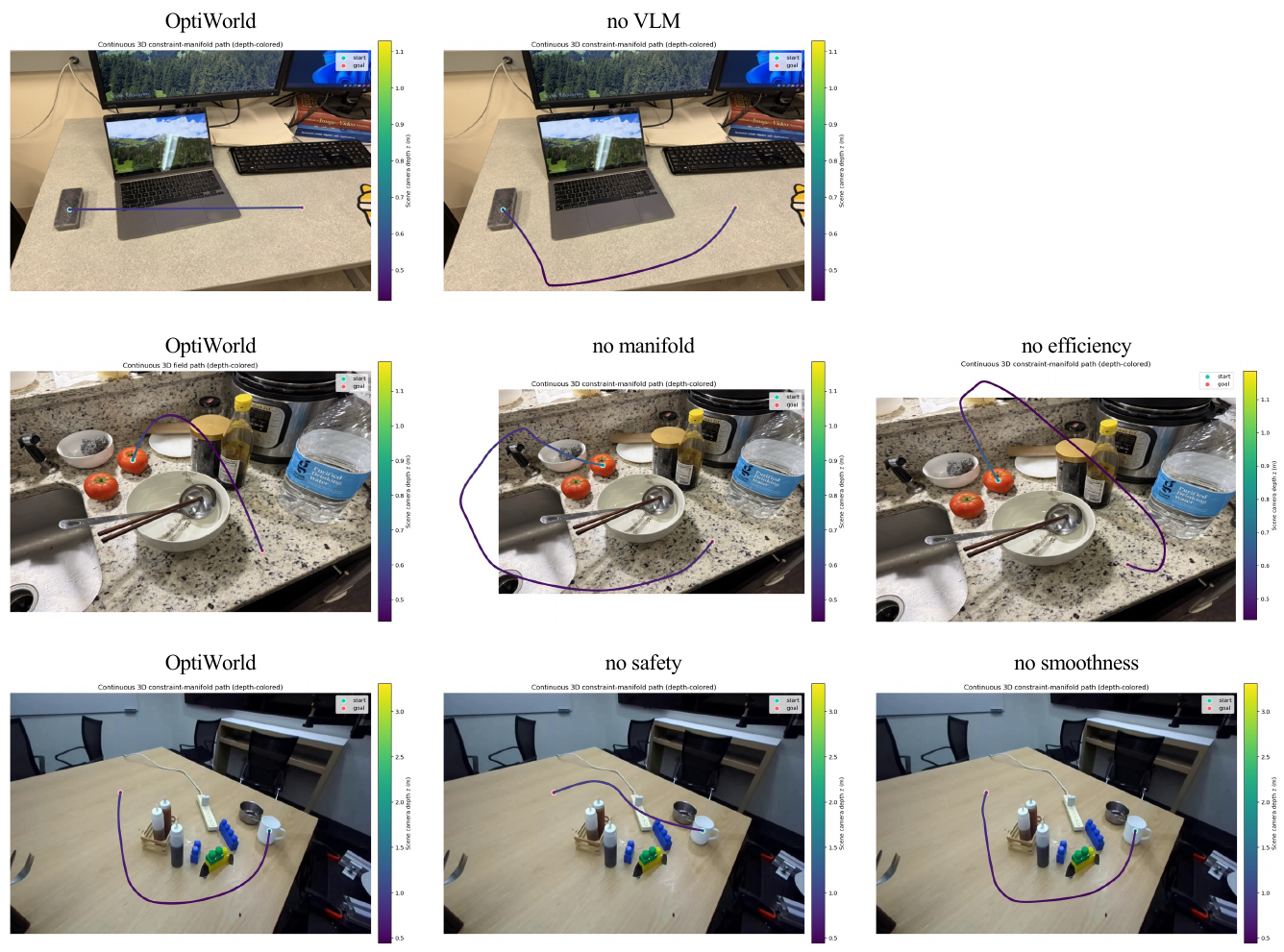}
    \caption{\textbf{Additional ablation study results.}}
    \label{fig:more_ablation}
\end{figure}

Figure~\ref{fig:more_ablation} provides additional visual examples for the ablation study. Removing VLM reasoning weakens task-dependent semantic constraints. Removing the manifold, safety, efficiency, or smoothness terms produces paths that are less safe, less direct, or less stable.

\section{More Results for Counterfactual Physics Generation}
\label{app:more_counterfactual}

\begin{figure}[h]
    \centering
    \includegraphics[width=\linewidth]{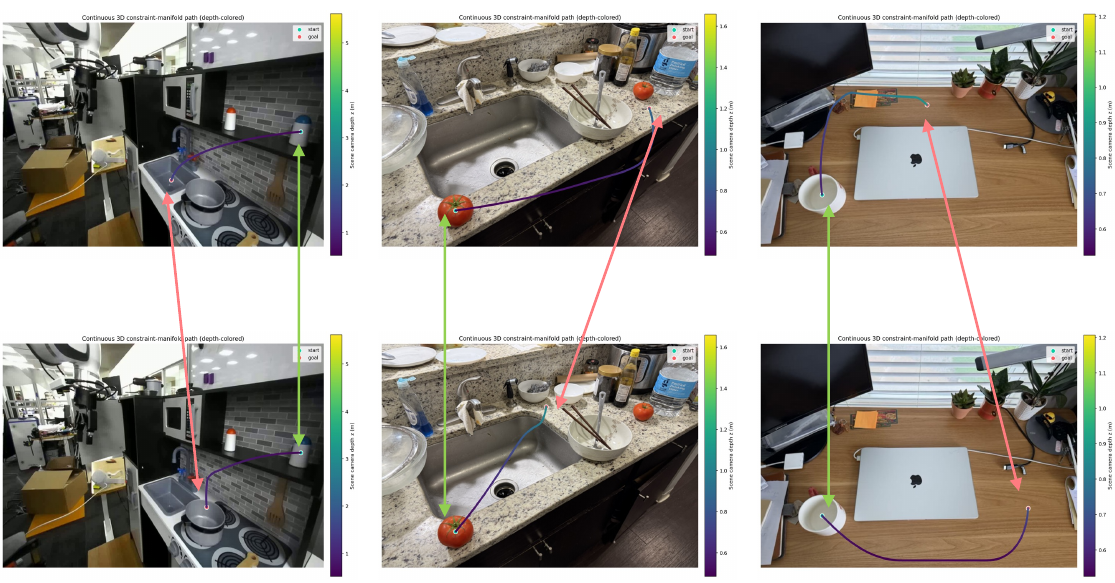}
    \caption{\textbf{Additional counterfactual results of goal control.} Different goal sets produce different optimized paths.}
    \label{fig:more_counter_multigoal}
\end{figure}

\begin{figure}[h]
    \centering
    \includegraphics[width=\linewidth]{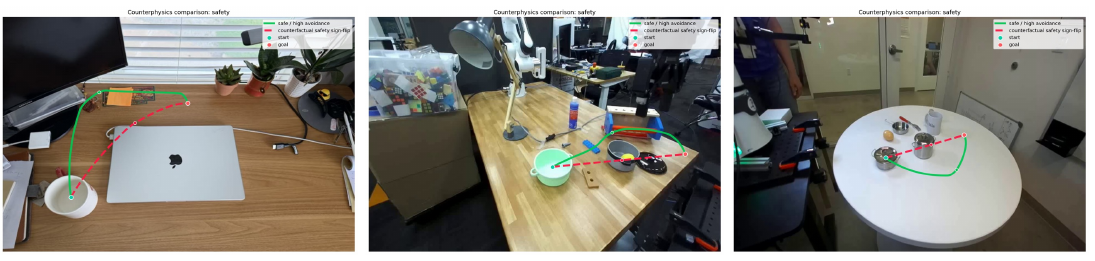}
    \caption{\textbf{Additional counterfactual results of safety control.} Lowering the safety constraint weight allows hazardous shortcuts, while the full planner avoids risky regions.}
    \label{fig:more_counter_safety}
\end{figure}

Figures~\ref{fig:more_counter_multigoal} and \ref{fig:more_counter_safety} show additional counterfactual generations produced by changing only the planner. Changing the goal gives different optimized trajectories for the same initial scene. Changing the safety weight changes whether the path avoids or enters risky regions, while the understanding and rendering modules remain fixed.

\section{Limitations and Broader Impacts}
\label{app:limitations}

OptiWorld improves motion planning before video generation, but the final video quality is still limited by the renderer. DaS can introduce artifacts, object deformation, or imperfect texture preservation, especially when the requested motion is large or the selected object is thin, reflective, or heavily occluded.

Another limitation is real-time use. The current system is designed as an offline inference pipeline. Future work should make the geometry estimation, semantic reasoning, planning, and rendering stages faster and more tightly integrated. A real-time version would be useful for interactive video world models and robot simulation.

\textbf{Broader Impacts.} This model could be used for generating high-quality content and educational videos; however, when misused without clear AI-generated content markers, it can produce highly convincing fake videos and thereby exacerbate the spread of misinformation.

\end{document}